\documentclass[10pt,twocolumn,letterpaper]{article}

\usepackage[pagenumbers]{cvpr} 

\usepackage{graphicx}
\usepackage{amsmath}
\usepackage{amssymb}
\usepackage{booktabs}
\usepackage{multirow}
\usepackage{algorithm}
\usepackage{algorithmic}
\usepackage{color}

\newcommand{\tablestyle}[2]{\setlength{\tabcolsep}{#1}\renewcommand{\arraystretch}{#2}\centering\footnotesize}

\newlength\savewidth

\usepackage[pagebackref,breaklinks,colorlinks]{hyperref}

\usepackage[capitalize]{cleveref}
\crefname{section}{Sec.}{Secs.}
\Crefname{section}{Section}{Sections}
\Crefname{table}{Table}{Tables}
\crefname{table}{Tab.}{Tabs.}

\def\cvprPaperID{1946} 
\def\confName{CVPR}
\def\confYear{2022}

\begin{document}

\title{Self-Supervised Class Incremental Learning}

\author{Zixuan Ni\\
Zhejiang University\\
{\tt\small zixuan2i@zju.edu.cn}
\and
Siliang Tang\\
Zhejiang University\\
{\tt\small siliang@zju.edu.cn}
\and
Yueting Zhuang\\
Zhejiang University\\
{\tt\small yzhuang@zju.edu.cn}
}
\maketitle

\begin{abstract}

Existing Class Incremental Learning (CIL) methods are based on a supervised classification framework sensitive to data labels. When updating them based on the new class data, they suffer from \textbf{catastrophic forgetting}: the model cannot discern old class data clearly from the new. In this paper, we explore the performance of \textbf{S}elf-\textbf{S}upervised representation learning in \textbf{C}lass \textbf{I}ncremental \textbf{L}earning (\textbf{SSCIL}) for the first time, which discards data labels and the model's classifiers. To comprehensively discuss the difference in performance between supervised and self-supervised methods in CIL, we set up three different class incremental schemes: \textbf{Random Class Scheme}, \textbf{Semantic Class Scheme}, and \textbf{Cluster Scheme}, to simulate various class incremental learning scenarios. Besides, we propose Linear Evaluation Protocol (LEP) and Generalization Evaluation Protocol (GEP) to metric the model's representation classification ability and generalization in CIL. Our experiments (on ImageNet-100 and ImageNet) show that SSCIL has better anti-forgetting ability and robustness than supervised strategies in CIL. 

To understand what alleviates the catastrophic forgetting in SSCIL, we study the major components of SSCIL and conclude that (1) the composition of different data augmentation improves the quality of the model's representation and the  \textit{Grayscale} operation reduces the system noise of data augmentation in SSCIL. (2) the projector, like a buffer, reduces unnecessary parameter updates of the model in SSCIL and increases the robustness of the model. Although the performance of SSCIL is significantly higher than supervised methods in CIL, there is still an apparent gap with joint learning. Our exploration gives a baseline of self-supervised class incremental learning on large-scale datasets and contributes some forward strategies for mitigating the catastrophic forgetting in CIL.


\end{abstract}

\section{Introduction}
Classes and concepts are increasing all the time in real life. However, the large-scale pre-training model based on supervised classification strategies cannot fit the new classes that never appear in the pre-training dataset \cite{krizhevsky2012imagenet}. The traditional method to solve this problem is that retrain the model by fusing new data with old datasets, but this is impractical in enterprises. Because retraining a large-scale pre-training model is time-consuming, labor-intensive, and costly, and has poor real-time performance. Therefore, it is best to update the model with new class data without the old dataset. We call this learning paradigm Class Incremental Learning (CIL) \cite{rebuffi2017icarl}. However, when we are directly fine-tuning the model with new class data, the model’s ability to distinguish old classes is rapidly declining. This phenomenon is called catastrophic forgetting \cite{goodfellow2013empirical}. Although many methods try to alleviate it \cite{rebuffi2017icarl,he2020incremental,hu2021distilling}, the performance is not satisfactory.

\begin{figure}[t]
	\centering       
	\includegraphics[width=8cm,height=4.3cm]{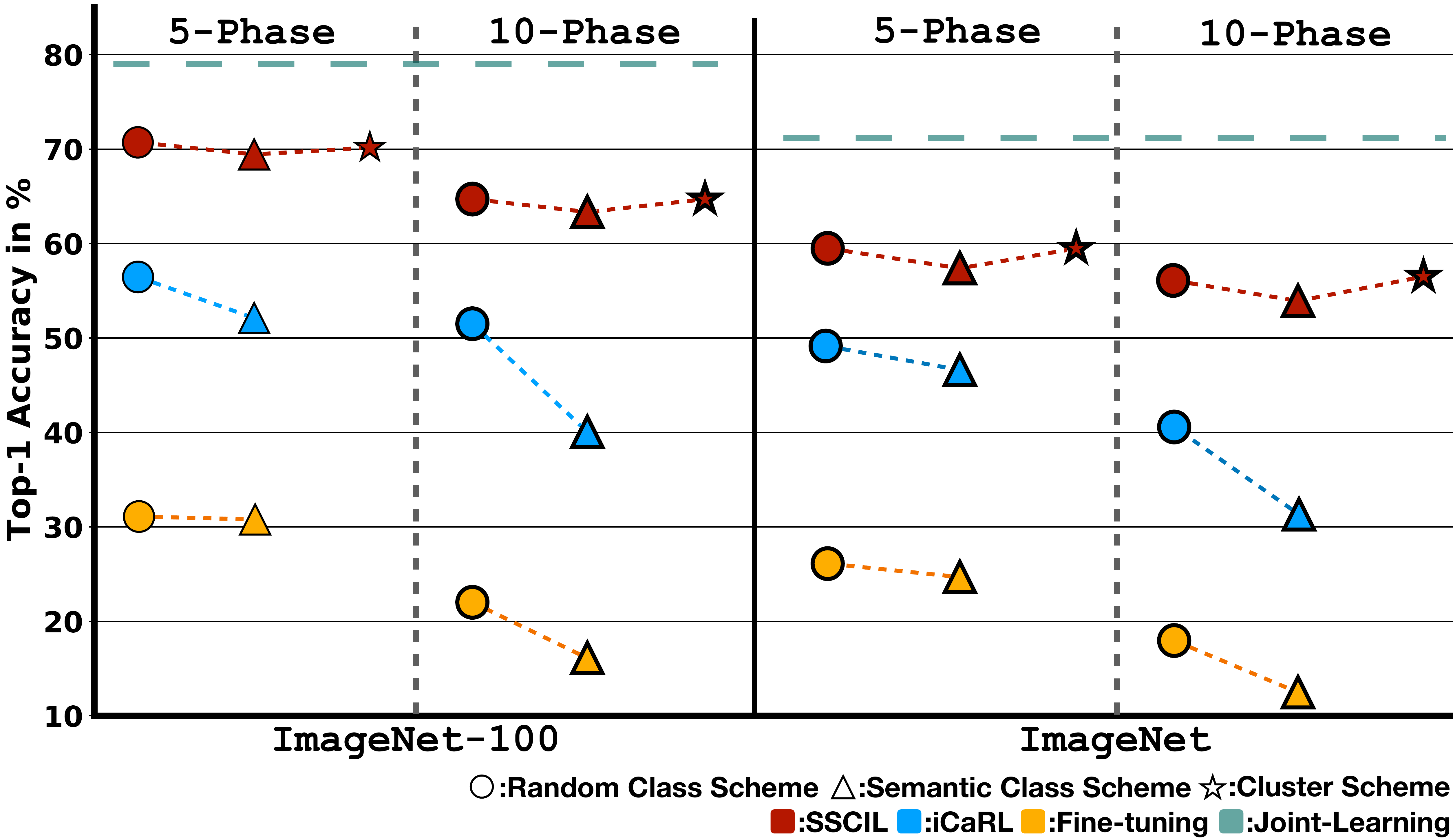}
	\caption{Comparing the final accuracy of the model in SSCIL with iCaRL and fine-tuning. The performance of SSCIL outperforms other methods in all situations.}
	\label{top-1result}
\end{figure}


Recently, \cite{ni2021alleviate} finds the reason for the catastrophic forgetting in the supervised fine-tuning model is that the linear separability of the model's representation decreases significantly with the addition of new class data progressively. They name this phenomenon \textit{Representation Forgetting}. We believe that the main reason for the representation forgetting is that the supervised model is susceptible to classes. In order to amplify the difference between classes, the supervised classifiers use conversion functions like SoftMax to extend the distance with different classes data in the classification layer. Because of this, the model only focuses on updating the parameters to fit the current dataset, which, however, destroys the previous model's representation space and leads to severe catastrophic forgetting.

In this paper, we discard the model's classifier and the label information to discuss the performance of \textbf{S}elf-\textbf{S}upervised \textbf{C}lass \textbf{I}ncremental \textbf{L}earning (SSCIL) on large-scale class datasets. The traditional scheme to simulate class incremental process is splitting dataset $D$ into $T$ sub-datasets based on random classes. And the number of classes in each sub-dataset is equal, and they are do not intersect. And we are then training the model in these sub-datasets orderly. We name this scheme \textbf{Random Class Scheme}. In order to discuss the difference between supervised and self-supervised methods in CIL more comprehensively, we add two different class incremental schemes:  \textbf{Semantic Class Scheme} and \textbf{Cluster Scheme} to simulate different class incremental sequences. The Semantic Class Scheme constructs sub-datasets based on different semantic information (like animal and vegetable). And the Cluster Scheme clusters the data in high-dimensional space through K-means and sets the training sequence based on different bunches. In order to evaluate the quality of the model's representation in the process of CIL, we propose Linear Evaluation Protocol (LEP) and Generalization Evaluation Protocol (GEP) to metric the model's classification ability and robustness at each training phase in CIL. We compare the LEP and GEP results of the SSCIL and supervised CIL on 5-phase and 10-phase ImageNet-100 and ImageNet with three class incremental schemes (Fig \ref{main_result}). In Figure \ref{top-1result}, based on linear evaluation protocol, we show the model's final accuracy in each situation. In Imagenet-100, SSCIL achieves 71.2\% top-1 accuracy in 5-phase and 69.8\% top-1 accuracy in 10-phase (the joint-learning of self-supervised is 79.8\%). In Imagenet, SSCIL achieves 65.03\% top-1 accuracy in 5 phase and 60.85\% top-1 accuracy in 10-phase (the joint-learning of self-supervised is 71.2\%). These results indicate that, unlike supervised methods that are sensitive to the shift of the training data domain, the performance of SSCIL is less affected by the different class incremental sequences. And SSCIL has a better anti-forgetting ability than other existing supervised methods in CIL.

In order to understand what alleviate catastrophic forgetting in SSCIL, we systematically analyze the influence of each component of the framework in SSCIL and find that:

\begin{itemize}
\item Although image distortion brings the system noise in CIL, we show that \textit{Grayscale} in data augmentation reduces this noise through the experiments. And because of this, data augmentation plays a role in improving the quality of the model representation in SSCIL. 

\item The projector like a buffer between the loss function and the model in the SSCIL. Using a projector at each training phase helps the model reduce unnecessary parameter updates and stabilizes the linear separability of the model representation. 
\end{itemize}

In summary, the main contributions of this work are three-fold. (1). For the first time, we systematically discuss the performance of self-supervised representation learning in the CIL from the representation level. (2). We study the role of each major component in SSCIL and discover the special functions of the data augmentation and the projector in SSCIL. (3). We give the baseline of the self-supervised representation learning in CIL and will provide all the training and testing codes in Github. Although SSCIL is better than supervised methods in large-scale datasets, catastrophic forgetting still exists (the gap between SSCIL and joint learning). 

\section{Related Work}
\subsection{Class Incremental Learning}

Except for fine-tuning method, there are many attempts to alleviate the catastrophic forgetting in the CIL from different directions \cite{li2017learning,rebuffi2017icarl,ni2021alleviate,hu2021distilling}. The most typical method iCaRL \cite{rebuffi2017icarl} adopts Rehearsal Strategy and Regularization Strategy. Rehearsal means saving old knowledg with a limited memory bank and replay them in training. The most common method is to select and preserve some old classes samples to the memory bank \cite{castro2018end,wu2019large,hu2021distilling}. A more challenging approach is pseudo-rehearsal with generative models \cite{he2018exemplar,zhao2021memory}. Regularization Strategies use the loss function to guide the update of the model's parameters to help the model save old knowledge. Recently, the most popular strategy is knowledge distillation \cite{hou2019learning,douillard2020podnet}. The iCaRL method uses the strategies: saving old classes samples and knowledge distillation. 


\subsection{Self-Supervised Learning}
Recently, Self-supervised contrastive learning \cite{oord2018representation,chen2020simple} has become the most popular method in representation learning. Different from the previous methods by predicting rotations \cite{gidaris2018unsupervised}, colorization \cite{larsson2016learning} and cluster assignments \cite{caron2018deep}, contrastive learning makes positive and negative samples from original data by data augmentation \cite{henaff2020data,bachman2019learning}. And then by optimizing the InfoNCE loss \cite{oord2018representation} to distinguish the positive samples from negative. 

Now, more and more companies focus on pre-training models with large-scale datasets \cite{radford2021learning}. Although large-scale datasets can cover many scenarios, when new classes data appear, the supervised model still cannot fit them well. Some works try to improve the supervised CIL by using unsupervised methods \cite{rao2019continual,gallardo2021self}. However, they still base on supervised labels, and the gap between them and iCaRL is tiny. Therefore, in this work, we explore the performance of self-supervised representation learning in CIL, which does not rely on label information.

\section{The Process of Self-Supervised Class Incremental Learning}
\subsection{Definition of Class Incremental Learning}
To simulate the process of CIL, we split the dataset $D$ into $N$ sub-datasets $\{D^{1},...,D^{N}\}$ and $D^{t}$ means the sub-dataset in incremental phase $t$ where $t=\{1,2,...,N\}$. 
When the model is training in the incremental phase $t$ , the previous sub-datasets $\{D^{1},...,D^{t-1}\}$ can no longer be used for training. 

\subsection{The Framework of SSCIL}
To explore the baseline of SSCIL, we follow the SimCLR \cite{chen2020simple} only using data augmentation $\mathcal{T}$, encoder $f(\cdot)$ , projector $g(\cdot)$, and contrastive loss $\mathcal{L}$ to construct the training framework of SSCIL. The data augmentation in SSCIL consist of four transformations. The first is \textit{RandomResizedCrop}, which interpolates the images to 224$\times$224. The second is \textit{random applied ColorJitter}, which transform the images' brightness, contrast, saturation and hue. The second to last is \textit{RandomGrayscale}. And finally, the \textit{GaussianBlur} and \textit{Solarization} be used. We use the same augmentation parameters as BYOL \cite{grill2020bootstrap}.

In training phase $t$, firstly, we randomly sample a minibatch of $K$ examples from sub-dataset $D^t$ and then use data augmentation $\mathcal{T}$ to distort them, getting $2N$ samples. Define the two augmented views $\tilde{x}^t_i$ and $\tilde{x}^t_j$ from the same sample $x^t$ as the positive pair and the rest are negative. Then, we get the representation $\tilde{h}^t$ from the encoder $f_t(\cdot)$ where $\tilde{h}^t=f_t(\tilde{x}^t)$. After that, through the projector $g_t(\cdot)$, we map the representation $\tilde{h}^t$ to the $\tilde{z}^t = g_t(\tilde{h}^t)$ which lies in the high-dimensional space. Finally, through the contrastive loss $\mathcal{L}$ to optimize the model.


\begin{equation}
\small
\mathcal{L} = -log\frac{\exp(s(\tilde{z}_i,\tilde{z}_j)/\tau)}{\exp(s(\tilde{z}_i,\tilde{z}_j)/\tau) + \sum_{k=1}^{k\neq j} \exp(s(\tilde{z}_i, \tilde{z}_k)/\tau)}
\label{eq:lossBarlow}
\end{equation}

where the $s$ is the similarity function $s(a,b) = a^Tb/||a||\cdot||b||$, which reflects the distance between two representations. And $\tau$ is a temperature super-parameter. When training in the next phase $t+1$, the framework will inherits the encoder $f_t(\cdot)$ and projector $g_t(\cdot)$ from the phase $t$ and update them in sub-dataset $D^{t+1}$. The flow diagram of the training process is in the supplementary material. 

\subsection{Evaluation Protocol in SSCIL}
Unlike the supervised CIL, there is not a classifier in the SSCIL. In order to evaluate the performance of the various methods at each training phase uniformly, we introduce the Linear Evaluation Protocol (LEP) \cite{chen2020simple} and the Generalization Evaluation Protocol (GEP) to metric the classification ability and robustness of the model's representation which indicates the upper bound of the model's performance.


\subsubsection{Linear Evaluation Protocol}
Linear Evaluation Protocol (LEP) has been used widely in representation learning \cite{wu2018unsupervised,grill2020bootstrap,zbontar2021barlow}. In there, we use the LEP to freeze the encoder network $f_t(\cdot)$ after the training in phase $t$ has finished, and use all of the sub-datasets $\{D^1,...,D^{t}\}$ have been trained previous to fit a linear classifier $M_t$ on top of the $f_t(\cdot)$. The test accuracy $LEP_t = M_t(f_t(D^t_{test}))$ indicates the model's ability to classify all known classes at the phase $t$, where the $D^t_{test}$ is the test set of the sub-datasets $\{D^1,...,D^{t}\}$. When the training phase $N$ has been finished, the gap between $LEP_N$ and joint-learning which train the joint-dataset $\{D^1,...,D^{N}\}$ directly is a proxy represents model's catastrophic forgetting. 


\subsubsection{Generalization Evaluation Protocol}
Besides LEP, we propose the Generalization Evaluation Protocol (GEP) to evaluate how much knowledge the model has obtained in CIL. Different from the linear evaluation method ($LEP$), the generalization evaluation ($GEP$) uses the full-classes dataset $D$ to fine-tune a linear classifier $M_t$. The test accuracy $GEP_t = M_t(f_t(D_{test})$ tells us how much knowledge the model has after each training phase finish, where the $D_{test}$ is the test set of the full-dataset $D$. If the GEP keeps rising in CIL, it means that the model learn more knowledge than it has forgotten in the CIL and vice versa. After the training phase $N$, the $GEP_N$ represents the linear separability of the model on the full-classes dataset $D$, which is the same as $LEP_N$. The Algorithm \ref{alg:SSCIL} summarizes the process of SSCIL.

\begin{algorithm}[t]
\caption{\label{alg:SSCIL} The training and evaluation process of SSCIL.}
\begin{algorithmic}
\FOR {$t$ in training phase $P\in \{1,...,N\}$} 
	\STATE \textbf{Input:} encoder $f_{t-1}$, projector $g_{t-1}$, sub-dataset $D^t$
	\STATE $f_t = f_{t-1}$, $g_t = g_{t-1}$
	\FOR {sampled minibatch $\{x^t\}_K$ in sub-dataset $D^t$}
		\STATE $\tilde{x}^t_i, ~ \tilde{x}^t_j = augment(x^t)$
		\STATE $\tilde{z}^t_i = g_t(\tilde{h}^t_i) = g_t(f_t(\tilde{x}^t_i))$
		\STATE $\tilde{z}^t_j = g_t(\tilde{h}^t_j) = g_t(f_t(\tilde{x}^t_j))$ 
		\STATE \textbf{define} $L_t = mean(\mathcal{L}(\{\tilde{z}^t_i\},\{\tilde{z}^t_j\}))$  \\
		$~~~~~$where i $\in$ $\{$$1,...,K$$\}$~and~j $\in$ $\{$$1,...,K$$\}$
		\STATE update $f_t$ , $g_t$ to minimize $L_t$ \\
	\ENDFOR
	\STATE Evaluate the $LEP_t$ and $GEP_t$ of the encoder $f_t(\cdot)$.
	\STATE \textbf{Output} $f_t$ , $g_t$
\ENDFOR
\end{algorithmic}
\end{algorithm}

\subsection{Three Different Class Incremental Schemes}
\subsubsection{Random Class Scheme}
The traditional strategy to simulate CIL is using \textbf{Random Class Scheme}, which splits dataset $D$ into $T$ sub-datasets based on random classes. And the number of classes in each sub-dataset is equal and they are do not intersect. Although this method can make each sub-data set completely different, the semantic space of each class-set is likely to have a large amount of overlap. For example, the class \textit{Labrador} in sub-dataset $D^i$ and the class \textit{Huskies} in sub-dataset $D^j$ contain some same semantic information even though they are different class names. When the model begins to train sub-dataset $D^j$ after $D^i$ in order,  the catastrophic forgetting of these two classes is concealed because of the similarity of semantic information between them. In order to discuss the different performance between SSCIL and supervised methods more comprehensively, we add two additional class incremental schemes: \textbf{Semantic Class Scheme} and \textbf{Cluster Scheme} to simulate various class incremental learning scenarios.

\subsubsection{Semantic Class Scheme} 
There are some similarities between different classes with some same semantics.  For example, although \textit{cobra} and \textit{rattlesnake} are various classes, they all belong to the snake family. We call this class relationship as WordNet \cite{denning1989science}. In order to avoid the overlap of semantics between classes in each training phase, we re-define each class according to its WordNet and split full-dataset $D$ into $N$ sub-datasets $\{D^{(1)},...,D^{(N)}\}$ where the samples in the same sub-dataset have similar semantic information while the samples in different sub-dataset only has a few overlapping in semantic space (such as animal, commodity, etc.). 

\subsubsection{Cluster Scheme}
Besides this, in many experiments \cite{selvaraju2017grad,radford2021learning}, we have repeatedly found that the criteria the model to classify images are not the same as that of humans. Because of this phenomenon, we formulate a Cluster Scheme. Firstly, using a supervised pre-trained encoder (ResNet50) to extract the representations from all samples and then using K-means to divide these representations into N subsets. Based on the image index of these representations in each subset, the samples are divided into N sub-datasets. This method mainly emphasizes the discreteness of each sub-dataset in a high-dimensional space. Because of this dividing method, each sub-dataset often includes many labels which are the same as other sub-dataset. That means there are many intersections between the classes of each sub-dataset.

\section{Analyze the Components of SSCIL}
We discuss the impact of the components of SSCIL on catastrophic forgetting and lay out our empirical conclusions, which is helpful for us to understand the performance of SSCIL.

\subsection{Default Setting}
We use ResNet-50 \cite{he2016deep} as the basic encoder, and setting the projector has three linear layers which with 8192 output units in each layer. Each linear layer carries a batch normalization layer and a ReLU nonlinearity except the last one. We set the batch size in SSCIL is 1080, the learning rate is 0.2, and the training epoch is 300 at each training phase. To simulate class incremental scenarios, we split the ImageNet-100 into 5 sub-datasets according to the three class incremental schemes respectively, which means 20 classes of new data are trained at each training phase.

\subsection{Grayscale operation resists the system noise of data augmentation in SSCIL}

There are so many works \cite{chen2020simple,zbontar2021barlow,he2020momentum} that prove that data augmentation is crucial for learning good model representation. According to the hypotheses and conclusions in \cite{wen2021toward}, we believe that images are composed of \textbf{sparse signals} and \textbf{spurious dense noise}. And the dense noise in samples is stronger than the sparse signal. The purpose of data augmentation is to remove the dense noise by distorting the data while highlighting the sparse features. 

\begin{figure}[b]
\centering
\begin{subfigure}[t]{0.48\linewidth}
\includegraphics[width=1.5in,height=1.in]{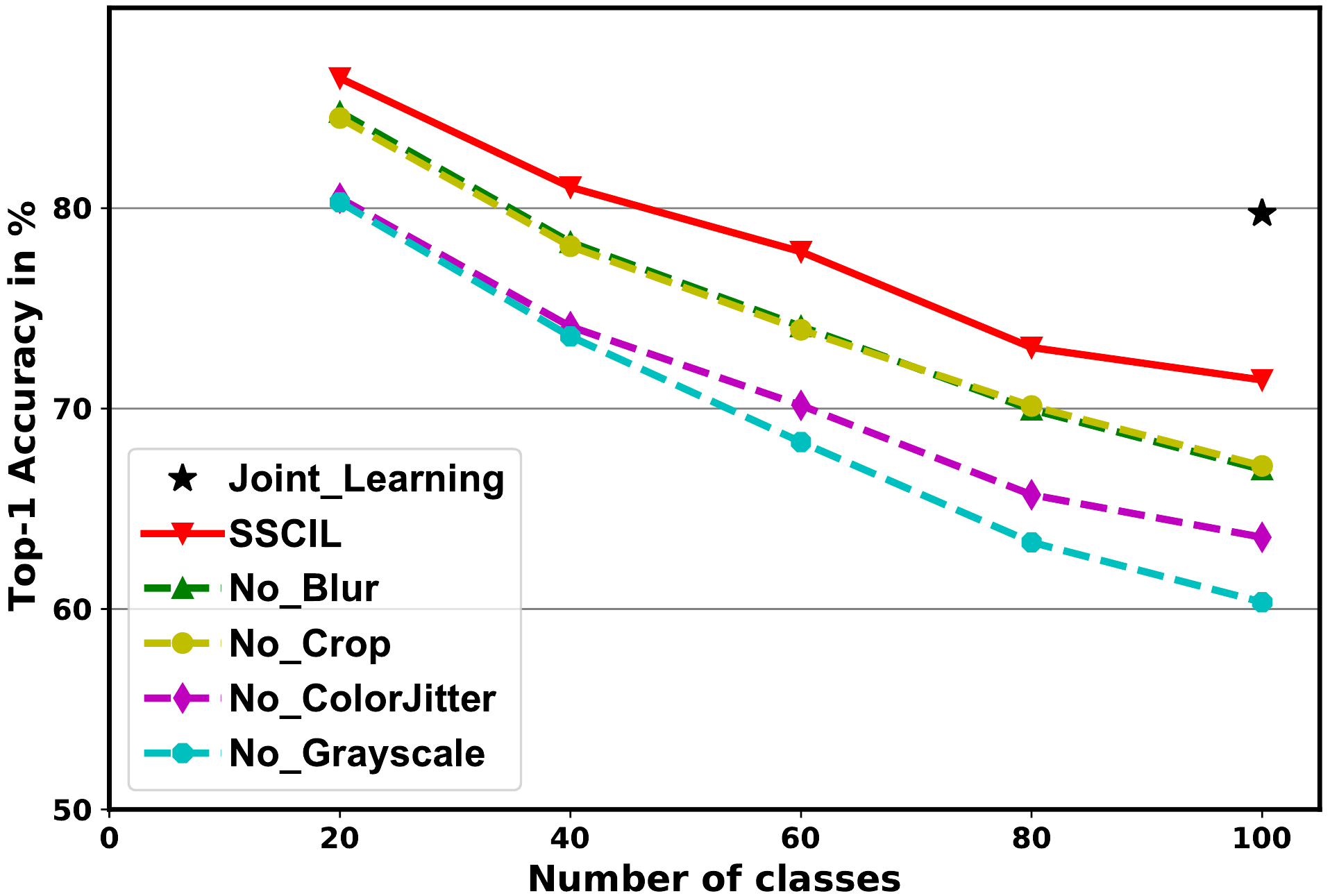}
\caption{}
\label{ablation_a}
\end{subfigure}
\begin{subfigure}[t]{0.48\linewidth}
\includegraphics[width=1.5in,height=1.0in]{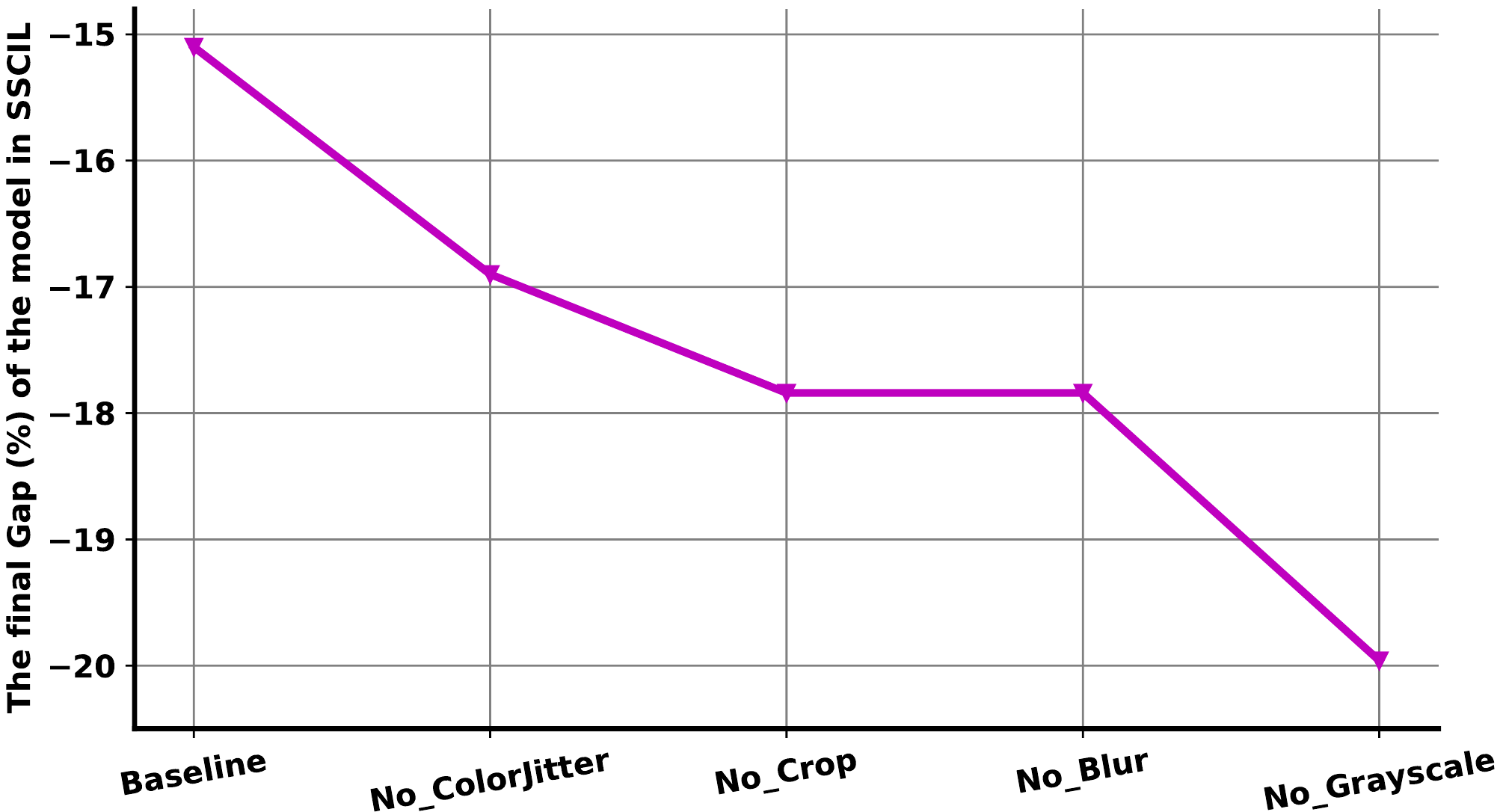}
\caption{}
\label{ablation_b}
\end{subfigure}

\caption{\textbf{a:} The LEP results when removing one of the data augmentation operations in SSCIL on the 5-phase Imagenet-100 with Random Class Scheme. \textbf{b:} The LEP gap between the result of the first training phase and the last for each ablation strategy.}
\label{aug_result}
\end{figure}

\begin{figure}[!h]
	\centering
	\begin{subfigure}[l]{0.49\linewidth}
	\centering
	\includegraphics[width=1.8in,height=1.57in]{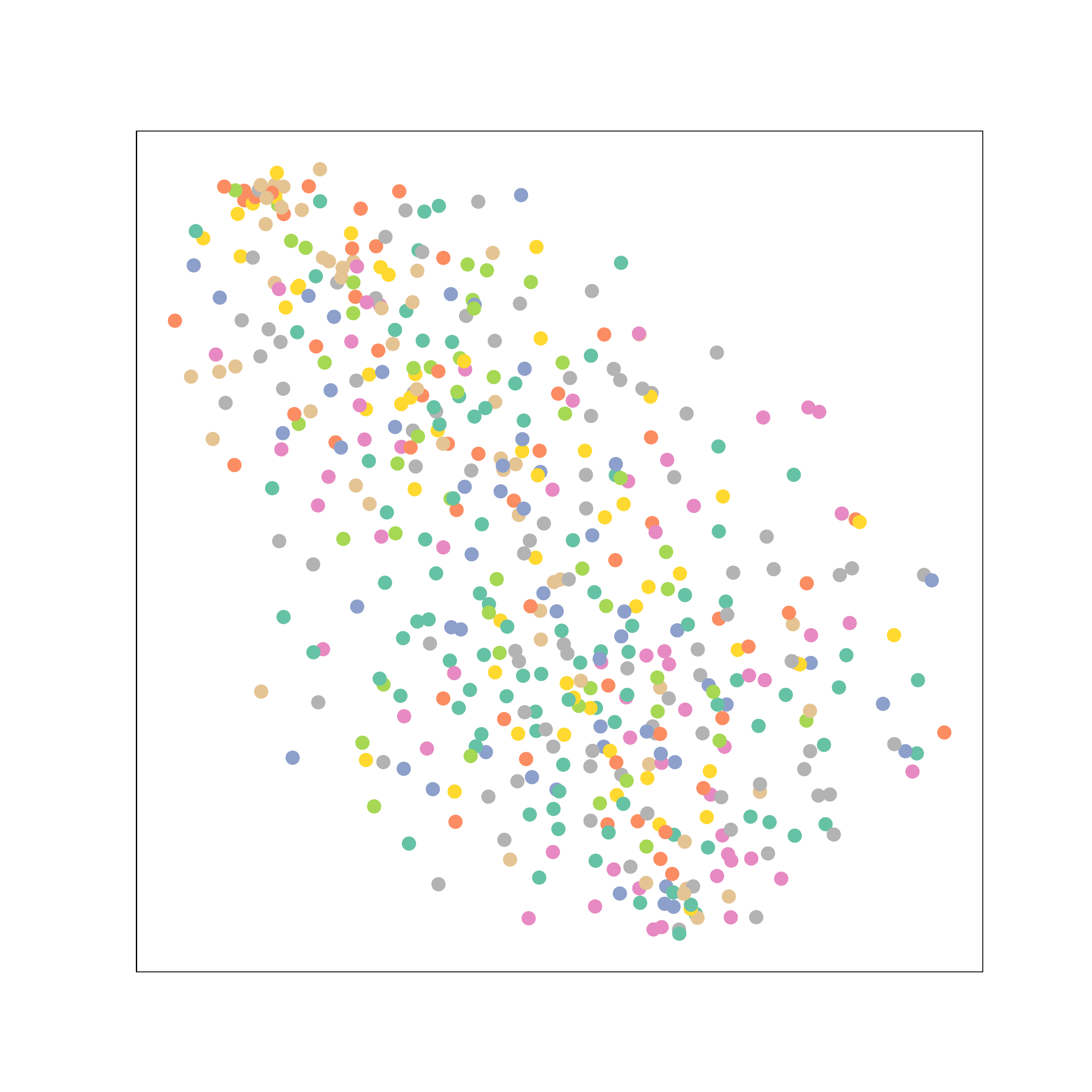}		\caption{Grayscale}
	\end{subfigure}
	\begin{subfigure}[l]{0.49\linewidth}
	\centering
	\includegraphics[width=1.8in,height=1.57in]{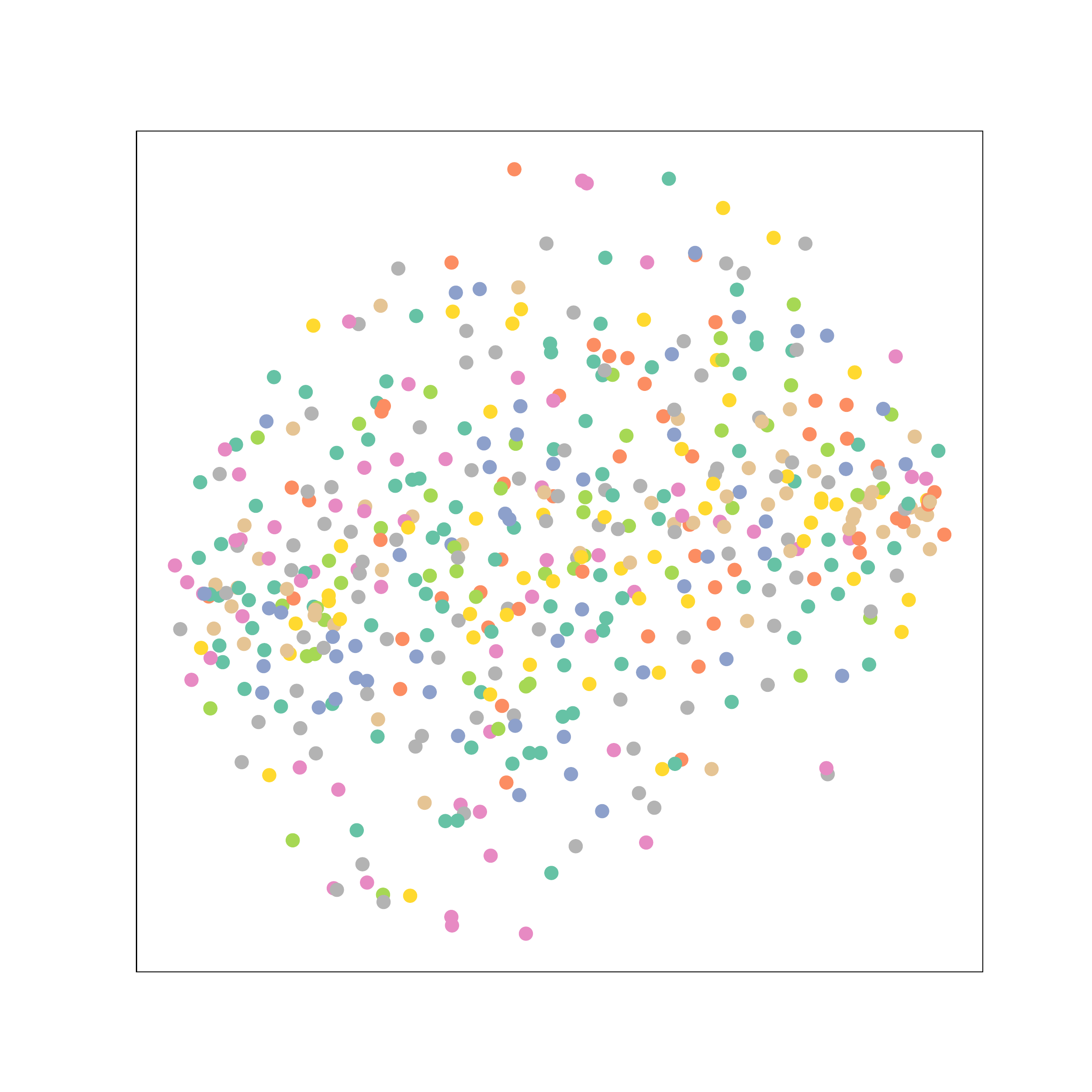}
	\caption{No \_ Grayscale}
	\end{subfigure}
	\centering
	\caption{The t-SNE of data distribution with or without grayscale operations.
	\textbf{(a)} Data distribution with grayscale operation. 
	\textbf{(b)} Data distribution without grayscale operation.}
	\label{grayscale_result}
\end{figure}

Here, we mainly discuss the impact of data augmentation in the process of SSCIL. We evaluate the model's performance when different data augmentation be removed on the 5-phase Imagenet-100 with Random Class Scheme (Fig \ref{ablation_a}). Comparing the models' results with different augmentation operations in the first training phase, we find that the LEP of the models drops from 86.5\% to 80.5\%. We obtain conclusions similar to the previous works that \textit{RandomGrayscale} and \textit{RandomColorJitter} have a significant impact on the representation quality of the model. However, when we compare the model's LEP performance at the end of the training phase, we find that the ultimate performance of the \textit{No\_Grayscale} is much lower than other ablation strategies, falling from the baseline 71\% to 60.34\%. Then, we compare the LEP gap between the models at the first training phase and the result of the final training phase for each ablation strategy in Fig \ref{ablation_b}. The \textit{No\_Grayscale} strategy causes most remarkable accuracy drop (from 80\% to 60\%). And the impact of \textit{ColorJitter} is relatively minimal. This phenomenon indicates that the \textit{Grayscale} operation in data augmentation slows down the catastrophic forgetting in SSCIL. Then, we use t-SNE \cite{maaten2008visualizing} to show the data distribution with or without grayscale operation in Fig.\ref{grayscale_result}. Compared with the data distribution without grayscale operation, we find that grayscale operation helps the data cluster to a uniform direction, and the dispersion is relatively reduced.

We think the reason is that: in CIL, due to the training of the model being broken down into multiple phases, the \textit{RandomColorJitter} operation in different training phases do not come from a common random seed system, which leads to a system noise that interferes with the model's learning of the sparse signals in data. Because of this, the model gradually fits the system noise with the rise of the training phase. And the Grayscale operation reduces this system noise by removing all RGB information from images, which like a common baseline for views be constructed by color augmentation combinations in different training phases. When training the self-supervised model by contrasting these views with their baselines, the sparse signal becomes apparent, guiding the model to learn a better representation. Because of this, Grayscale operation eliminates the system noise that comes from data augmentation, which helps prevent catastrophic forgetting in SSCIL.

\subsection{The projector reduces unnecessary parameter updates of the model in SSCIL}
When training a self-supervised model on a large-scale dataset, \cite{chen2020simple} shows that the width and depth of the projector affect the representation quality of the model. However, in SSCIL, we find that the width and depth of the projector head don't significantly affect the linear separability of the model's representation. What's more, whether or not to use a projector significantly affects the representation quality of the model in SSCIL. 

\begin{figure}[h]
\begin{minipage}[l]{0.65\linewidth}
\includegraphics[width=2.2in,height=1.5in]{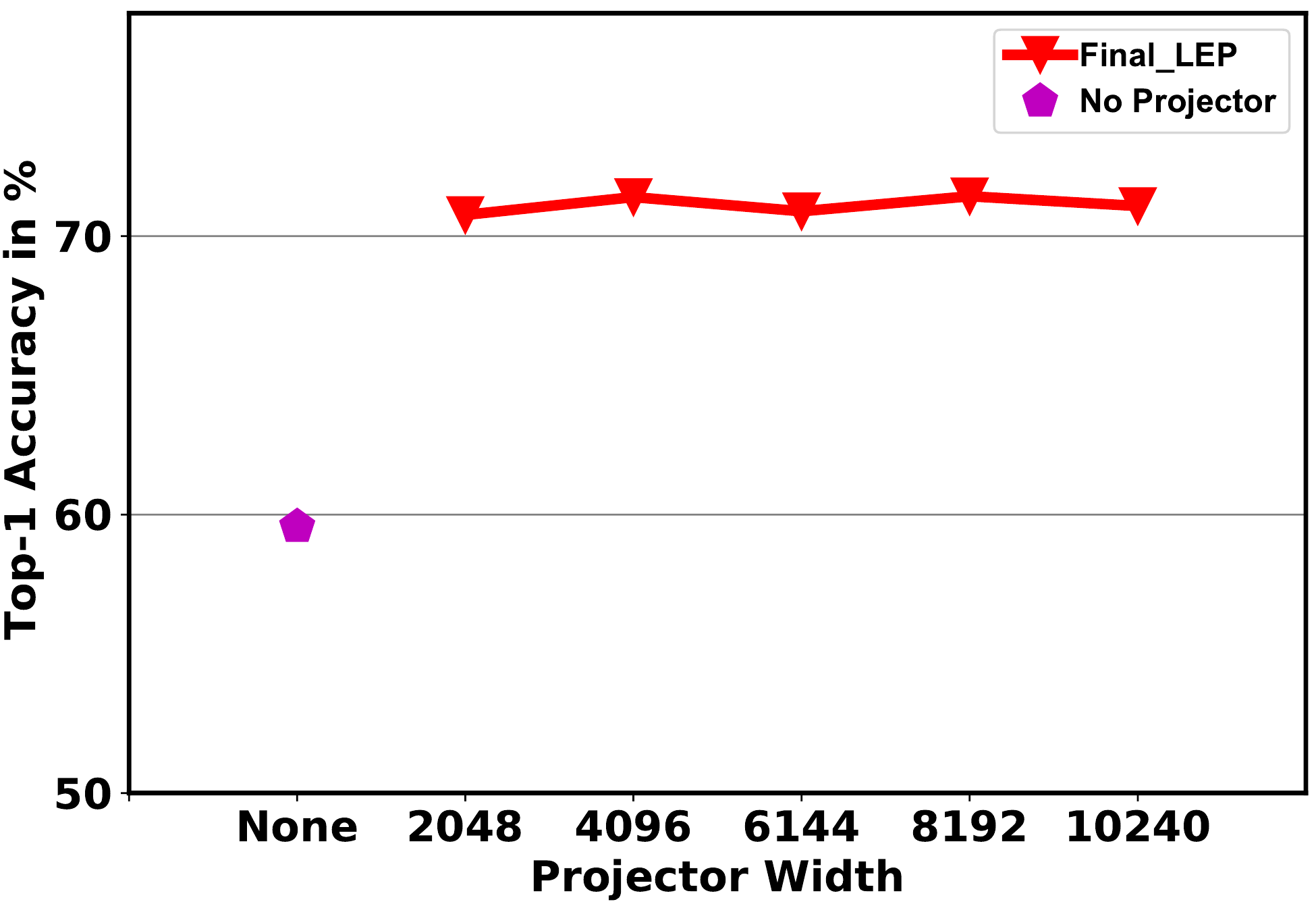}
\end{minipage}
\begin{minipage}[r]{0.33\linewidth}
  \small
  \tablestyle{3.0pt}{1.4}
  \begin{tabular}{c|c}
  \textbf{Depth} & {\textbf{LEP}} \\ 
  
  \hline
  {\textbf{8192x1}} & 69.64\\
  {\textbf{8192x2}} & 70.26\\
  {\textbf{8192x3}} & \textbf{71.43}\\
  {\textbf{8192x4}} & 70.7\\
  {\textbf{8192x5}} & 69.8\\
  \end{tabular}
  \vspace{-0.3em}
\end{minipage}
\caption{The final result of SSCIL with different projector structures on the 5-phase Imagenet-100 with Random Class Scheme. \textbf{Left plot:} The accuracy (\%) of the SSCIL with different width projector head (all is 3 linear layers) after the final phase. And the \textit{None} represents \textit{No Projector} in SSCIL. \textbf{Right table:} Accuracy (\%) of the different depth projector head on the final phase.}
\label{structure_result}
\end{figure}

In Figure \ref{structure_result}, we show the performance of the projector with different depths and widths on the 5-phase Imagenet-100 with Random Class Scheme. The width result is based on the three linear layers in a projector. From the left plot in Figure \ref{structure_result}, we can find that increasing the dimension of each linear layer in the projector does not significantly improve the linear separability (LEP) of the SSCIL after the last training phase is completed. The final LEP is all-around 71\%. Comparing the results of different depth projectors from the right table, we get conclusions similar to those of varying width projectors; The depth of the projector will also not promote the catastrophic forgetting of the model. 

\begin{figure*}[t]
\begin{subfigure}[t]{0.495\linewidth}
\centering
\includegraphics[width=1.6in,height=1.2in]{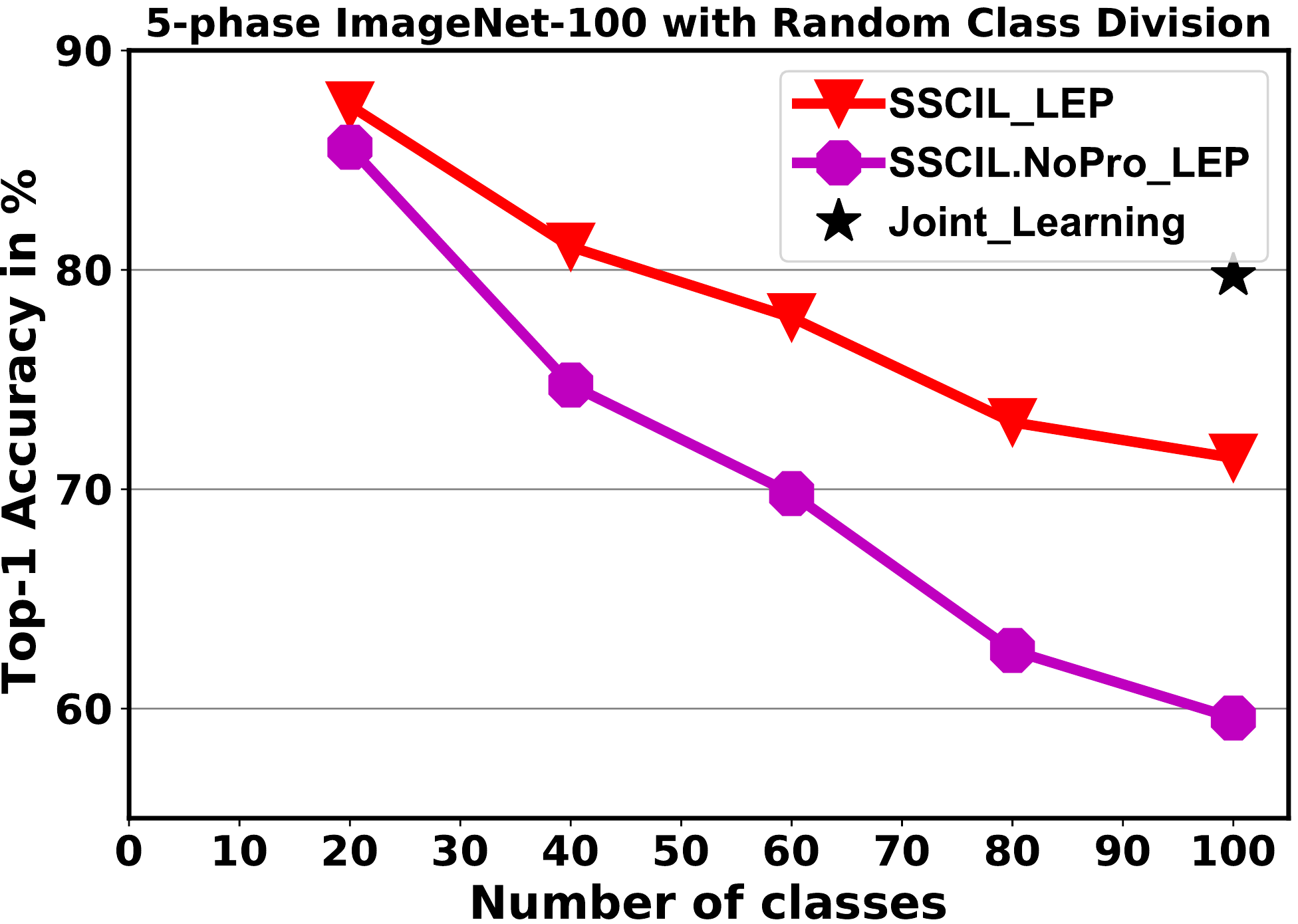}
\includegraphics[width=1.6in,height=1.2in]{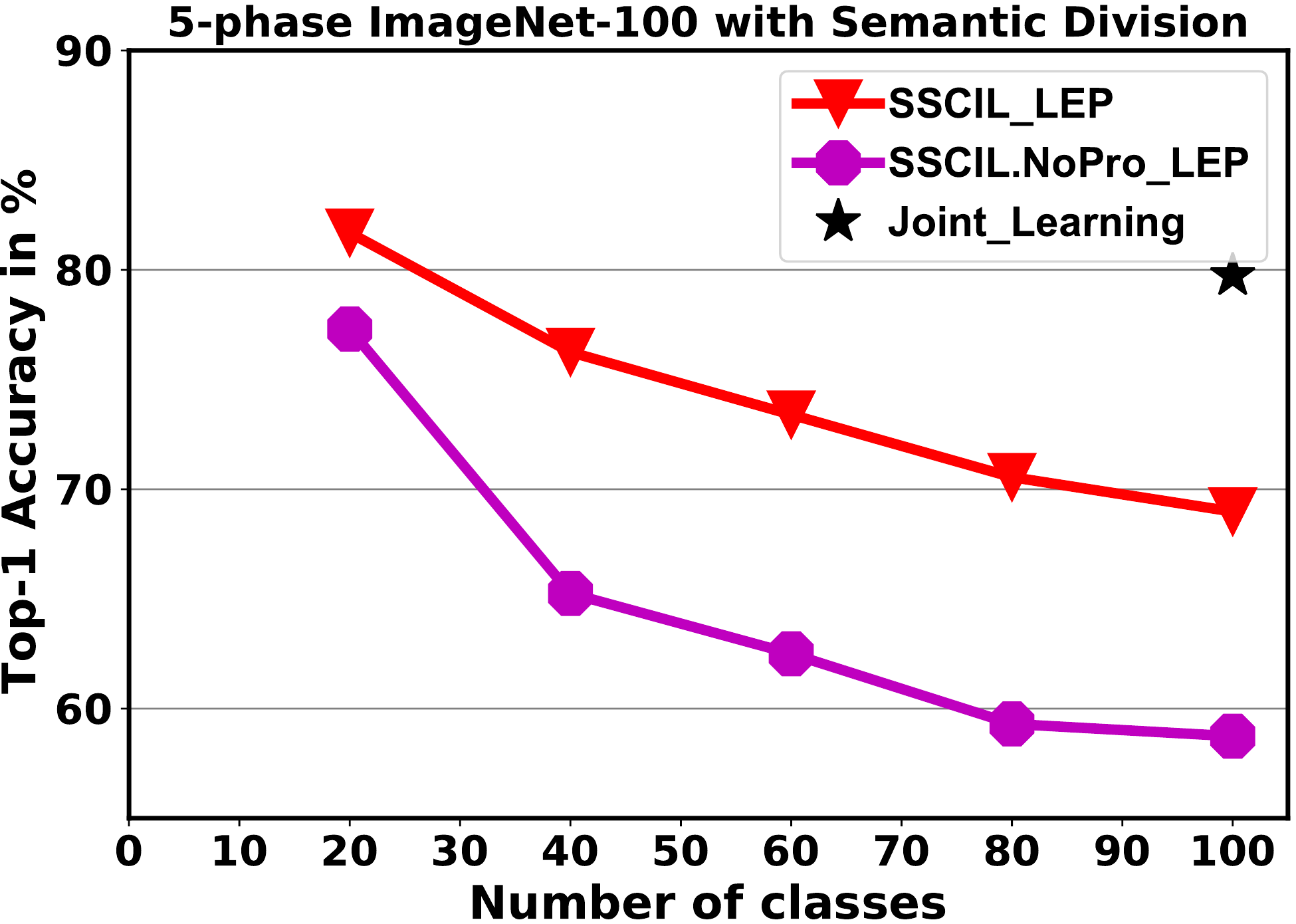}
\end{subfigure}
\begin{subfigure}[t]{0.495\linewidth}
\centering
\includegraphics[width=1.6in,height=1.2in]{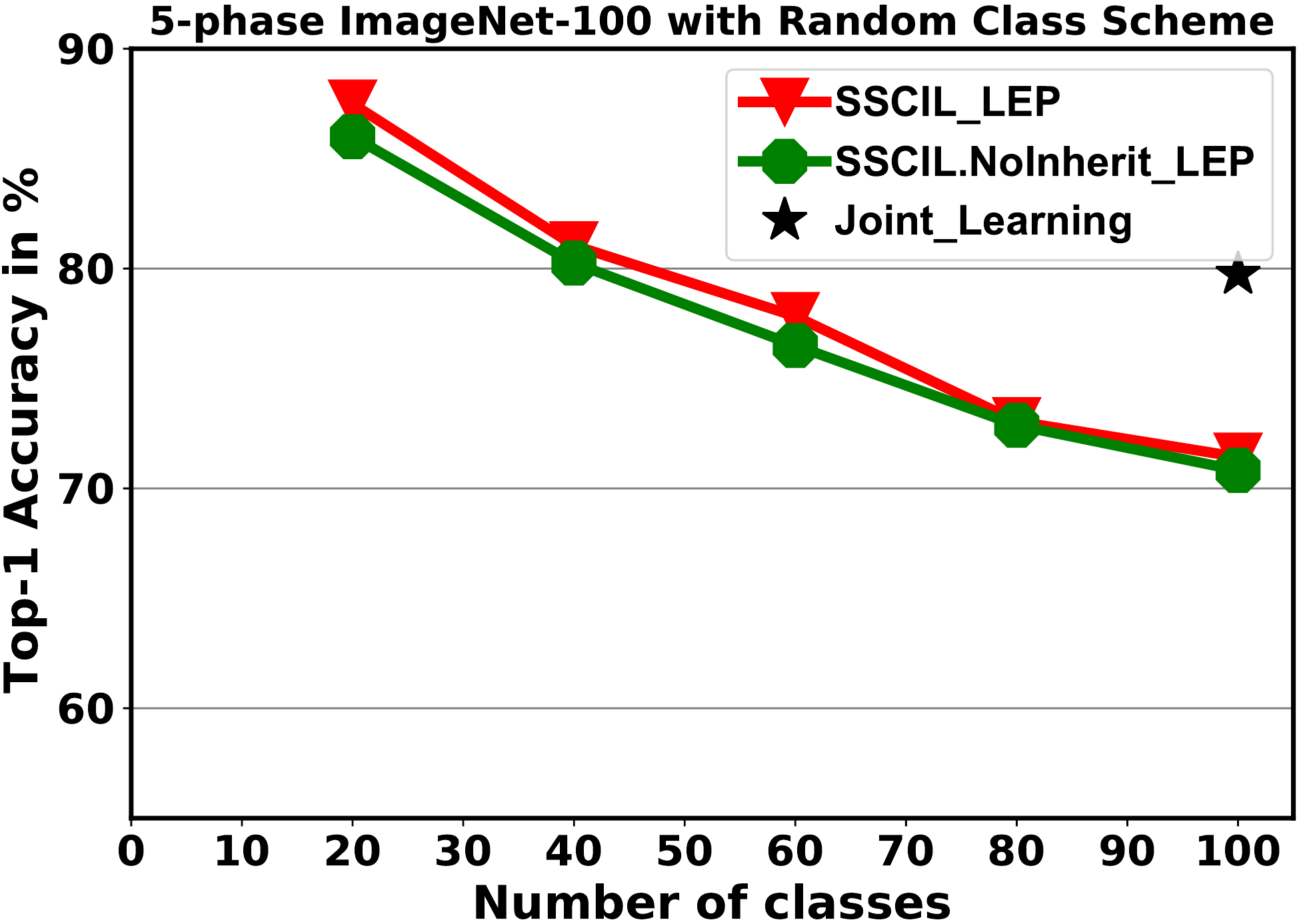}
\includegraphics[width=1.6in,height=1.2in]{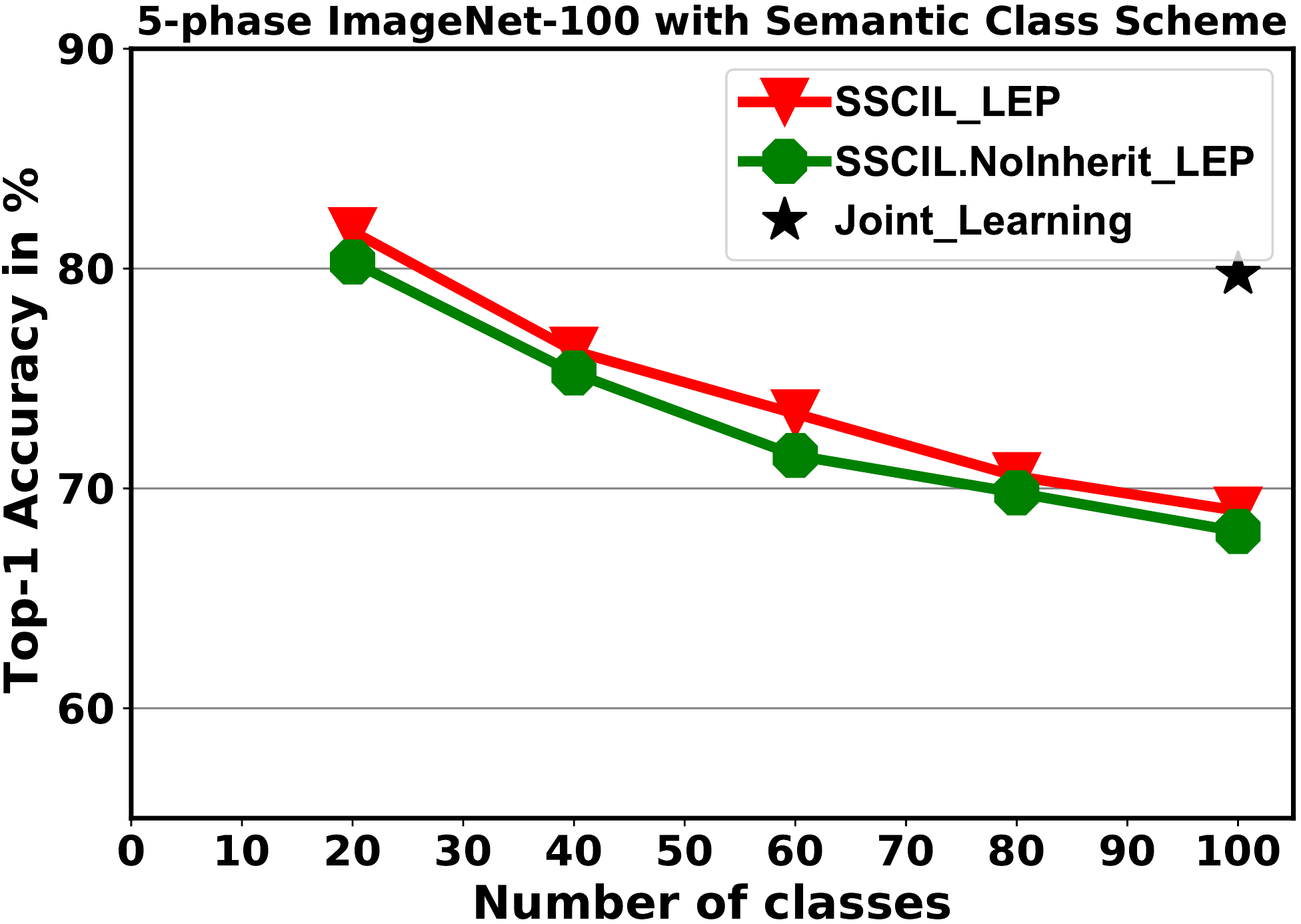}
\end{subfigure}
\caption{\textbf{Left plot:} Comparing the LEP of the SSCIL at each training phase with or without a projector on the 5-phase ImageNet-100 with Random Class Division and Semantic Division. With the rise of the training phase, the LEP of No Projector is gradually pulled apart by baseline. \textbf{Right plot:} The LEP result of the SSCIL at each training phase inheriting or not inheriting projector's parameter on the 5-phase ImageNet-100 with Random Class Scheme and Semantic Class Scheme.}
\vspace{-.3em}
\label{initial_result}
\end{figure*}

However, when we abandon the projector from the framework of SSCIL, we found that catastrophic forgetting is evident in the representation space of the model. The final LEP of the model drops from 71\% to 59\% (in Figure \ref{structure_result}). In the left plot of Fig.\ref{initial_result}, we show the model's LEP results at each phase on the 5-phase ImageNet-100 with Random Class Scheme and Semantic Class Scheme with or without a projector (3 linear layers). We can find that when a projector is applied, the LEP of the SSCIL in the first training phase rises from 85.59\% to 87.49\% on Random Class Scheme and 77.3\% to 81.69\% on Semantic Class Scheme. This indicates that a projector can improve the model's representation quality which is consistent with other works \cite{chen2020simple,zbontar2021barlow}. When we compare the results in the later training phase, the \textit{No\_Projector} strategy makes the linear separability of the model's representation (LEP) drop significantly than the baseline. Besides this, we discuss the impact of the learning strategy that uses the projector in each training phase but does not inherit the projector's parameters from the last training phase on the LEP accuracy of the SSCIL (the right plot of Fig.\ref{initial_result}). From the result, we can find that \textit{No\_Inherit} strategy just affects the quality of the model's representation slightly (-1\% $\sim$ -2\%) at each training phase.

\begin{figure}[h]
\begin{subfigure}[t]{1\linewidth}
\centering
\includegraphics[width=0.63in,height=.8in]{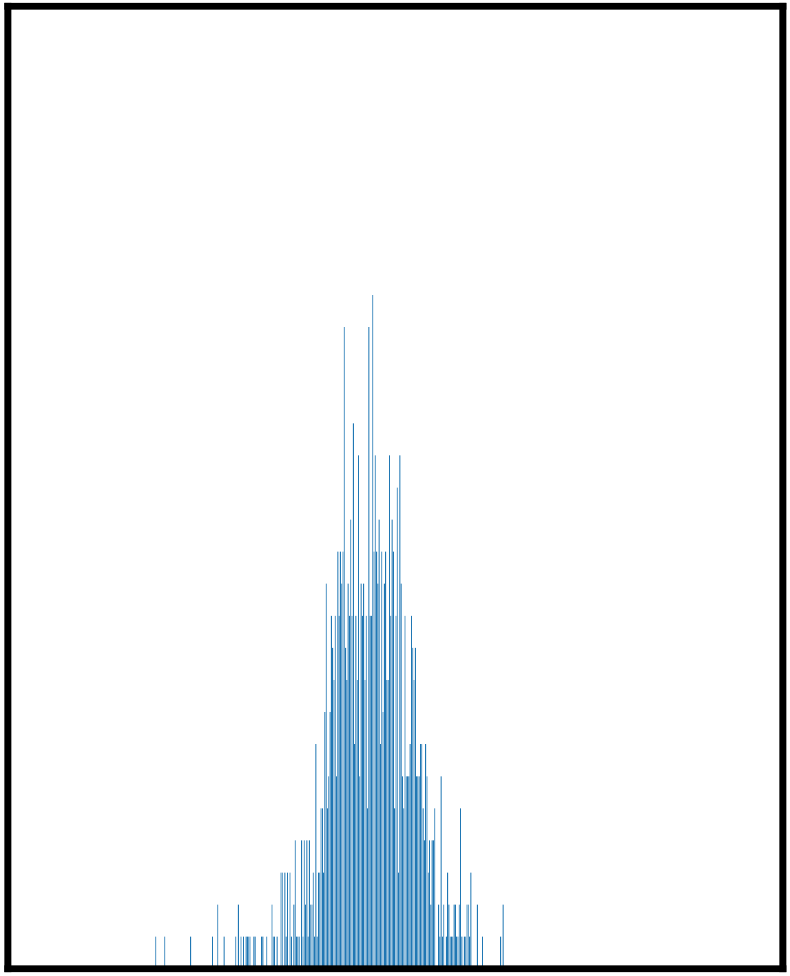}
\includegraphics[width=0.63in,height=.8in]{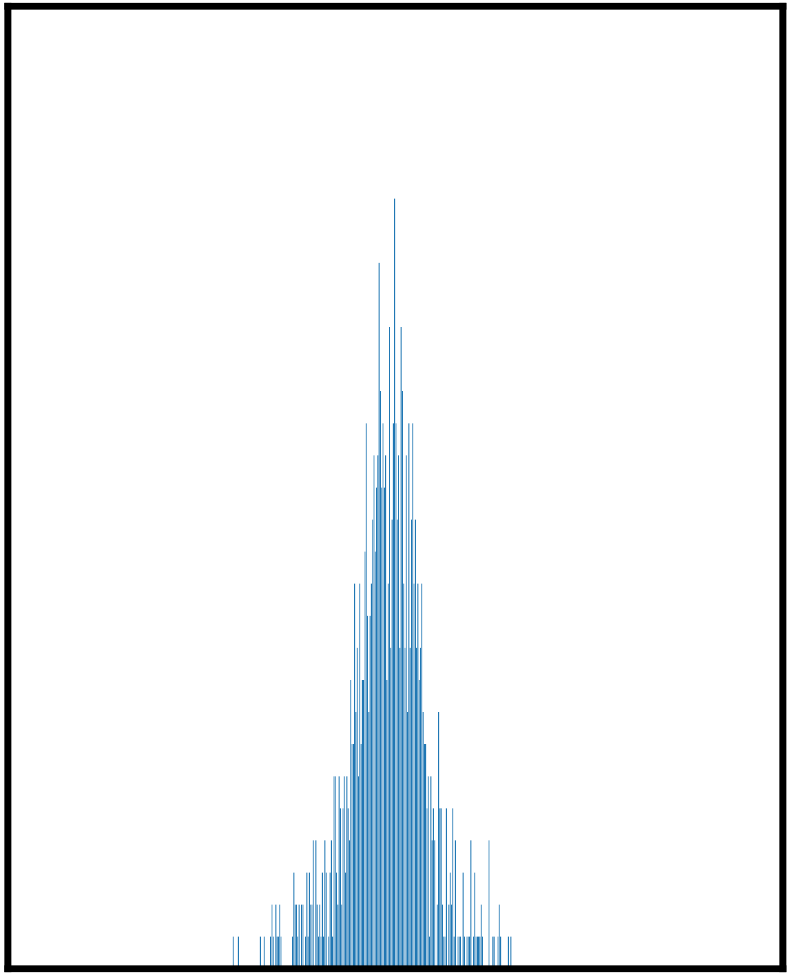}
\includegraphics[width=0.63in,height=.8in]{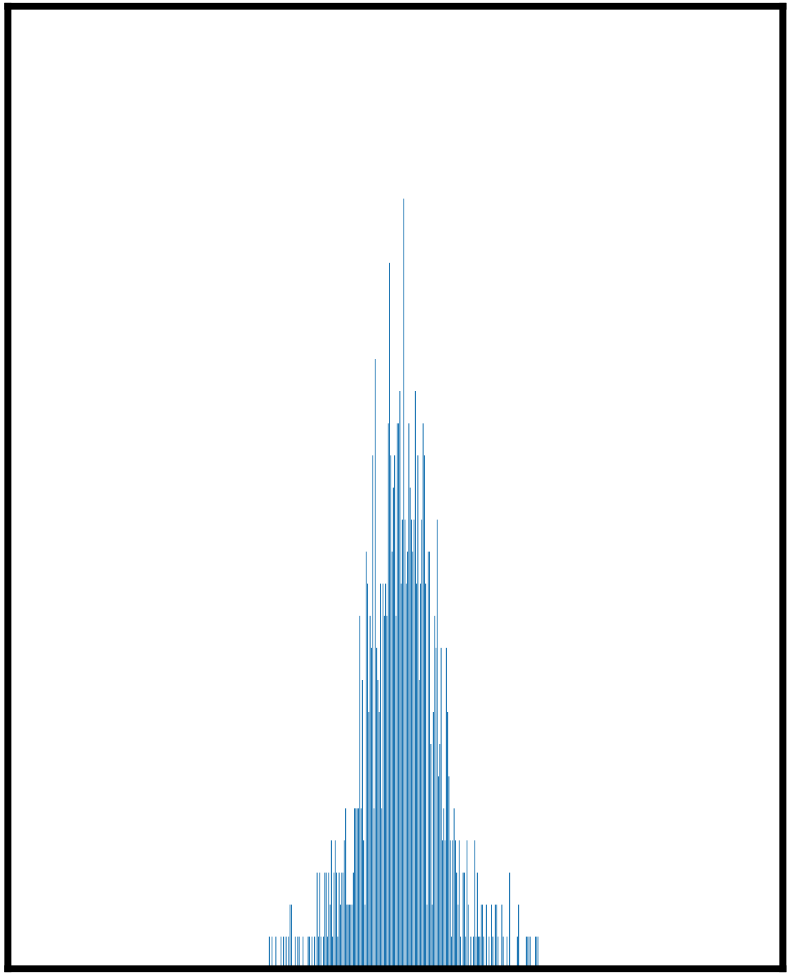}
\includegraphics[width=0.63in,height=.8in]{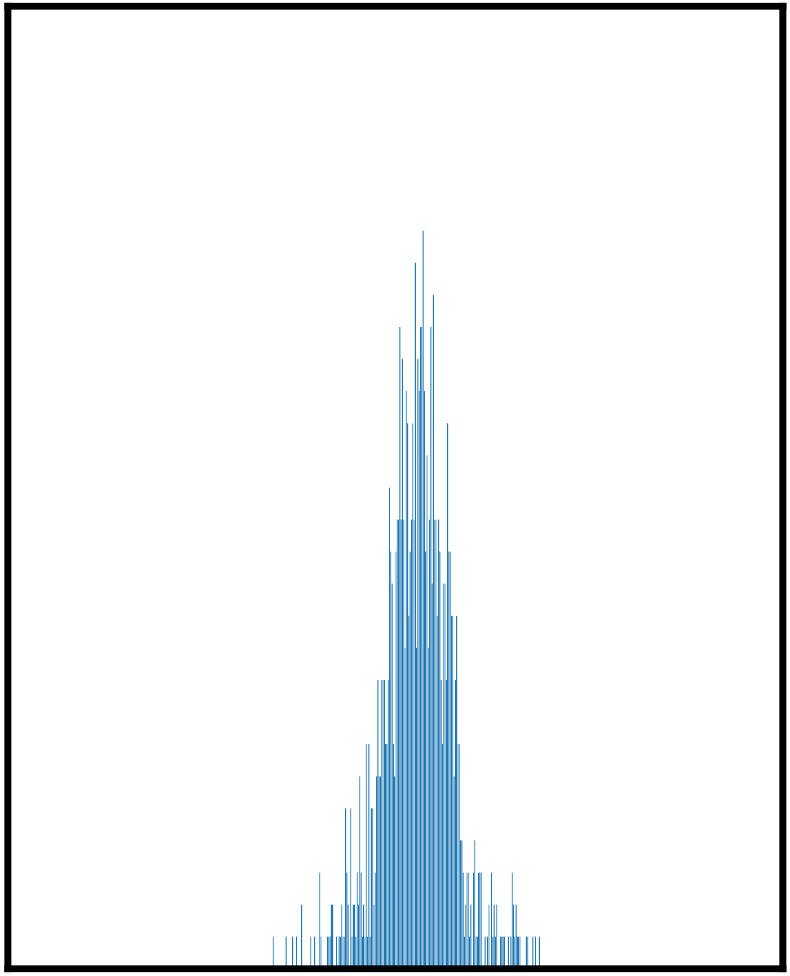}
\includegraphics[width=0.63in,height=.8in]{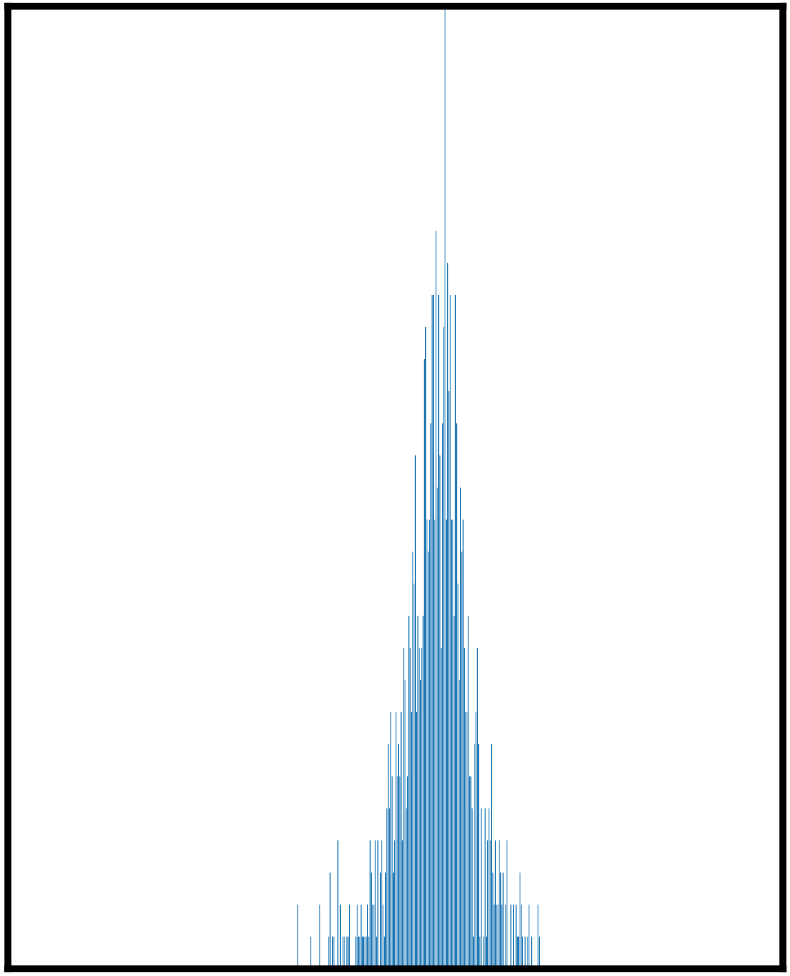}
\caption{The representation distance distribution on Encoder.}
\label{subresult_a}
\end{subfigure}

\begin{subfigure}[t]{1\linewidth}
\centering
\includegraphics[width=0.63in,height=.8in]{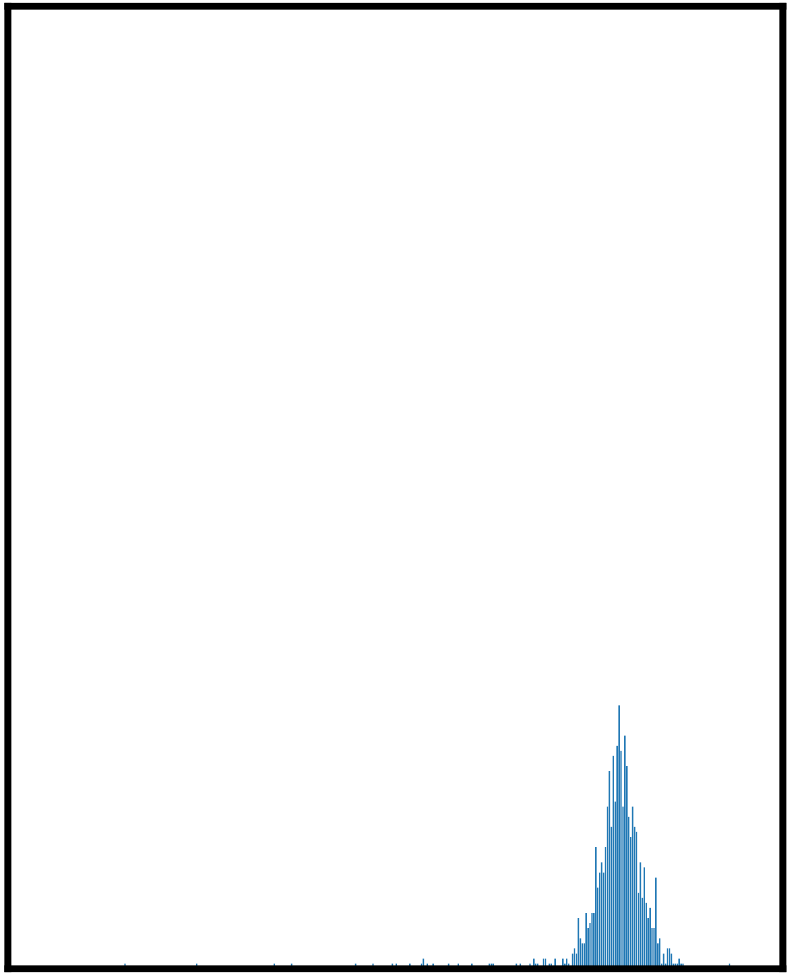}
\includegraphics[width=0.63in,height=.8in]{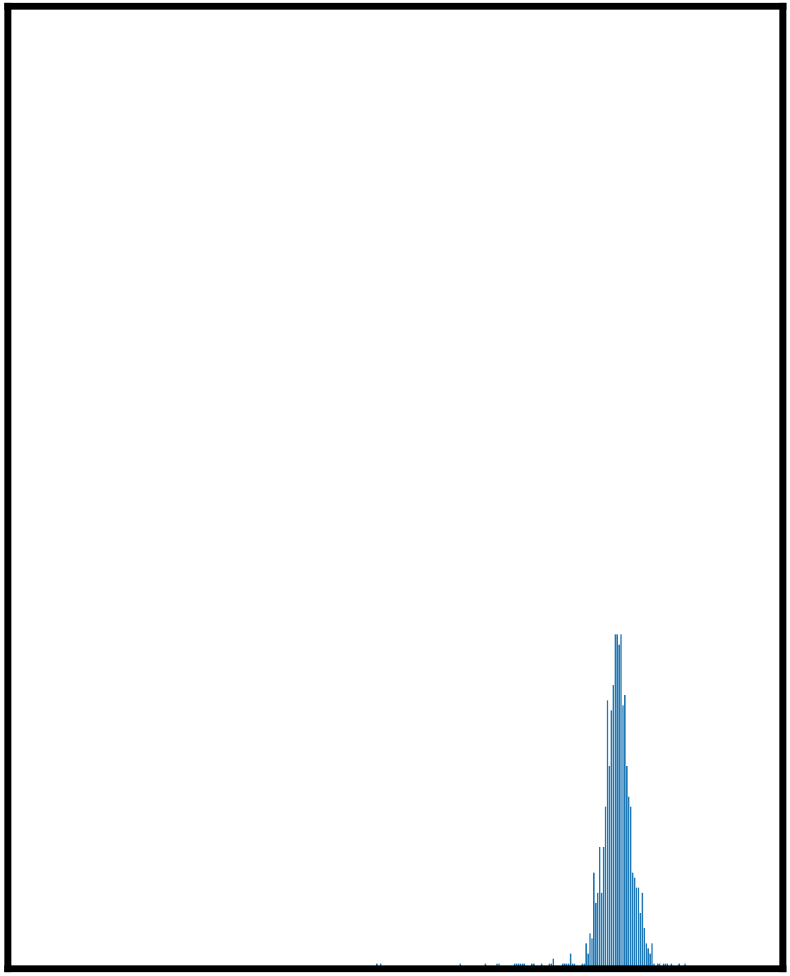}
\includegraphics[width=0.63in,height=.8in]{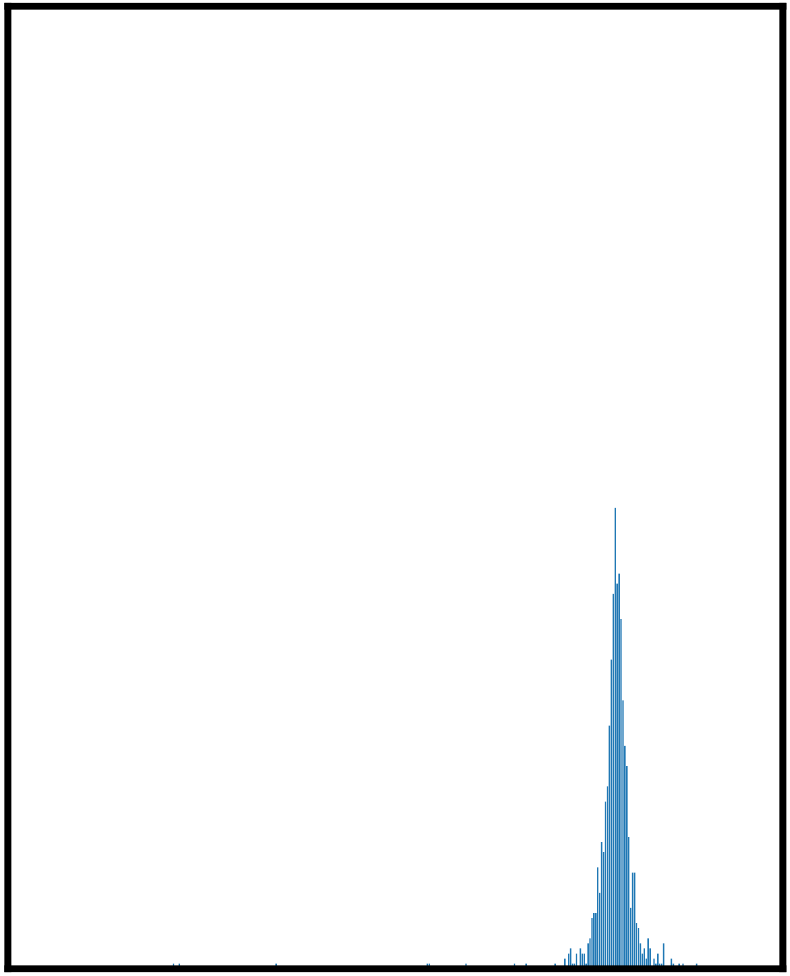}
\includegraphics[width=0.63in,height=.8in]{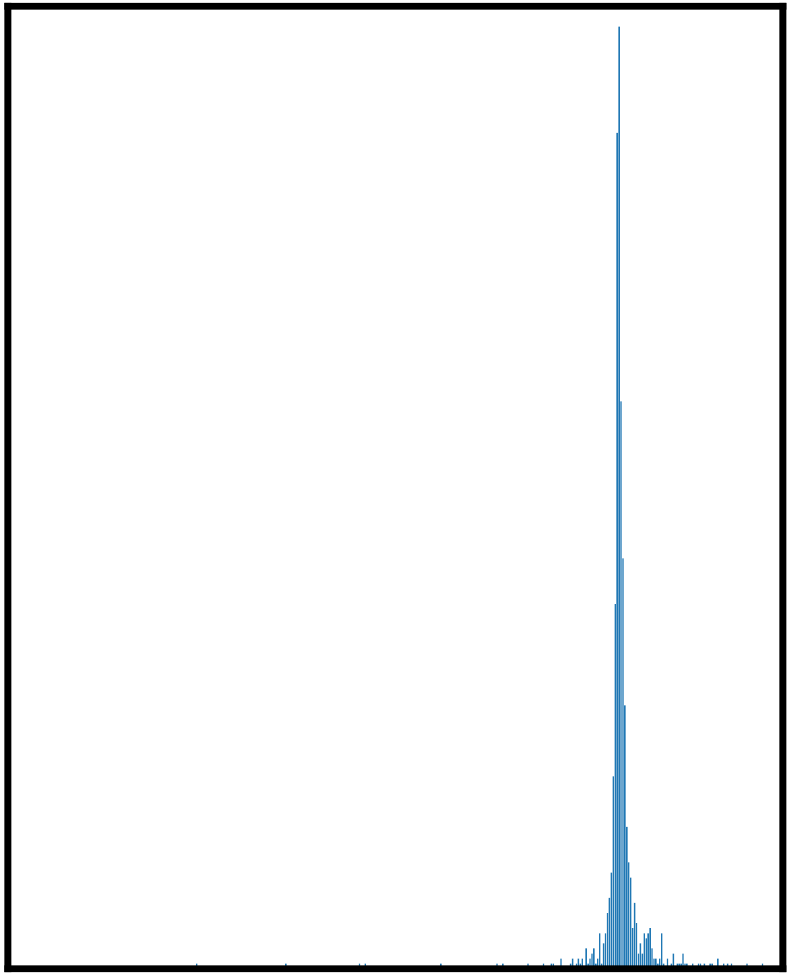}
\includegraphics[width=0.63in,height=.8in]{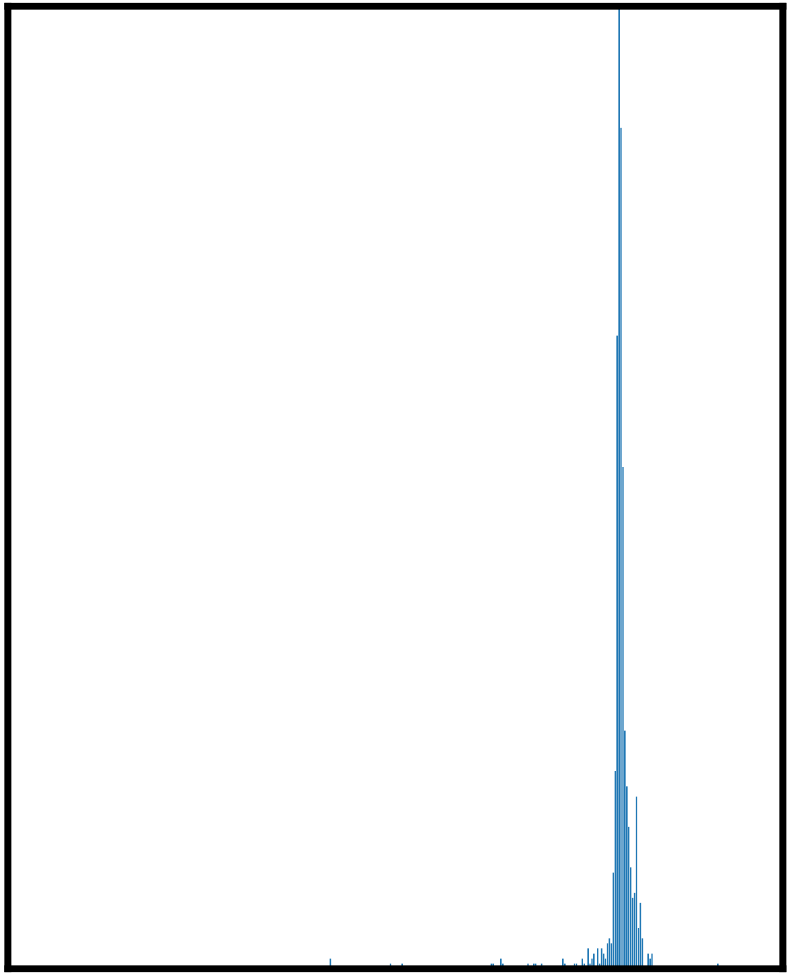}
\caption{The representation distance distribution on Projector.}
\end{subfigure}
\caption{Histograms of representation distance between each other. From the left to right, the training phases of the SSCIL gradually rises. }

\vspace{-1em}
\label{distribution_result}
\end{figure}

All of this proves that projector structure alleviates catastrophic forgetting of SSCIL observably. Fig.\ref{distribution_result} shows the representation distance distribution of the first training phase's test samples on Encoder and Projector from the first training phase to the last. Observing the sub-results \ref{subresult_a}, we find that the representation distance distribution of the test sample after Encoder does not shift significantly with the increase of the training phases. However, the representation distance distribution on the projector becomes  sharper and sharper as the training phase rises. We think the reason is that projector maps the representation relationship into a high-dimensional space. And when updating the model's representation in the training phase, the projector acts as a buffer, alleviating the direct influence of the loss function on the encoder. So the unnecessary update of the model's parameter is reduced, and the robustness is enhanced. 

\section{Comparison with Supervisied CIL Methods}

\subsection{Experiment Setting}
We compare the performance of SSCIL with supervised CIL methods (iCaRL \cite{rebuffi2017icarl} and fine-tuning) on ImageNet-100 \cite{tian2019contrastive} and ImageNet \cite{russakovsky2015imagenet}. 
As the most classic and still competitive method in the supervised CIL, the iCaRL has a simple model architecture and some typical anti-forgetting strategies (using regularization function and saving old samples), which is easy to reproduce and compare. The fine-tuning method, as the baseline of various CIL methods, clearly shows the catastrophic forgetting of the model in the CIL process. We use ResNet-50 \cite{he2016deep} as the basic encoder for all methods and set the initial learning rate of iCaRL and fine-tuning to 0.2. 

\begin{figure*}[!t]
	\centering
	
	\begin{subfigure}[t]{1\linewidth}
	\includegraphics[width=6.7in,height=2.85in]{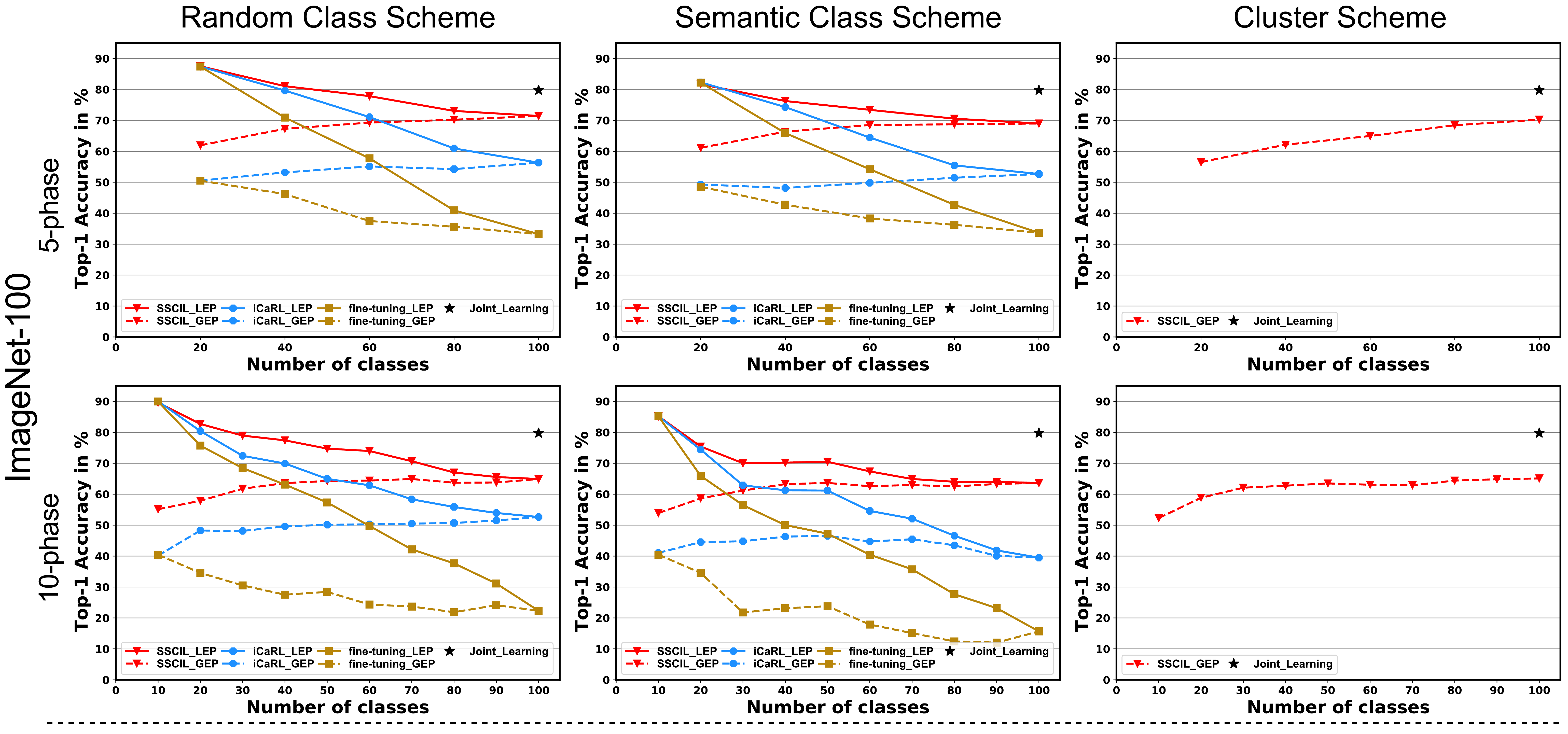}
    \end{subfigure}%
    
	\begin{subfigure}[t]{1\linewidth}
	\includegraphics[width=6.7in,height=2.85in]{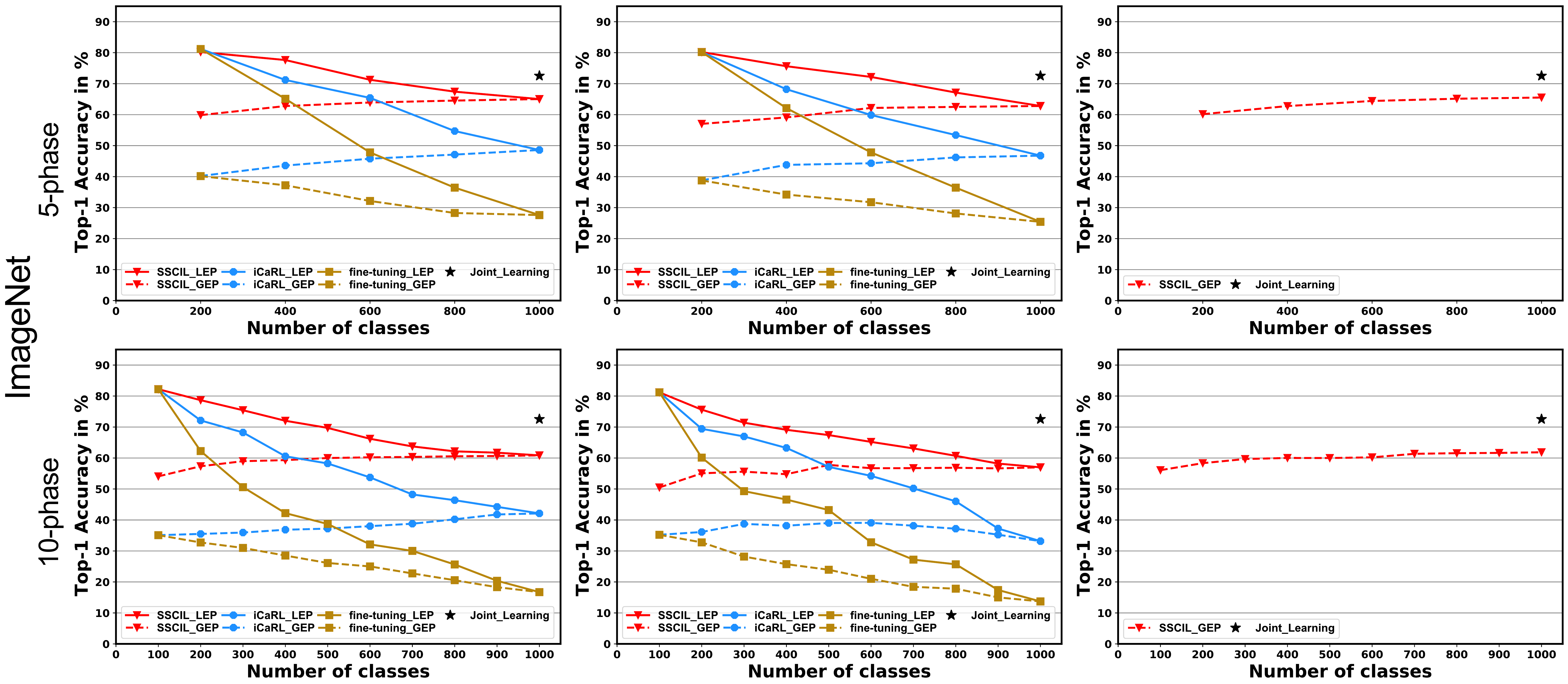}
	\end{subfigure}%
	
	\centering
	\caption{Illustrate the LEP and GEP performance of the method SSCIL, iCaRL and supervised fine-tuning on 5-phases and 10-phases ImageNet-100 and ImageNet with three different class incremental schemes. From the left to right are Random Class Scheme, Semantic Class Scheme, and Cluster Scheme. From the top to bottom are 5-phases and 10-phases ImageNet-100 and ImageNet. It can be seen from the results that the top-1 accuracy of SSCIL in any case surpasses the supervised methods, which show that SSCIL has a better anti-forgetting ability than supervised methods. The results of GEP reflect that SSCIL can improve its representation quality robustly, which ignores the shift of data domain in the CIL.}
	
	\label{main_result}
\end{figure*}

To simulate various class incremental scenarios, we split the ImageNet-100 and ImageNet into 5 and 10 sub-datasets according to the different class incremental schemes respectively, which means 20 and 10 classes of new data are trained at each training phase on ImageNet-100 or 200 and 100 classes of new data are trained at each training phase on ImageNet. All results are shown in Figure \ref{main_result}. Since the Cluster Scheme does not rely on the classes of samples to divide the dataset, each sub-dataset contains uneven class samples. So the validation set cannot be constructed in each sub-dataset, and the model's LEP cannot be evaluated in this strategy. So we just show the GEP results of the SSCIL at each training phase in the Cluster Scheme.

\subsection{Comparing with Different Training Phases.} 
From figure \ref{main_result}, we can find that no matter which class incremental scheme (from the left to right) is applied, on the ImageNet-100 dataset, with the increase of training phase, the LEP of SSCIL (solid line) is higher than other supervised CIL. The final LEP of SSCIL is 70\% and 65\% on 5-phase and 10-phase ImageNet-100, while the iCaRL is 58\% and 50\%, respectively. The trend of the gap between the SSCIL with supervised CIL on ImageNet is the same as ImageNet-100. This phenomenon shows that SSCIL has a better ability to resist catastrophic forgetting than supervised methods. Besides, when we compare the model's GEP results (dotted line) on different datasets, we find that the GEP of SSCIL increases with the rise of the training phase (for example, it grows from 60\% to 70\% on the 5-phase Imagenet-100). In contrast, the result of iCaRL fluctuates or does not change when the training phase is increasing. And the fine-tuning method even has a significant decline. All of this indicates that the knowledge SSCIL possesses rises during the process of CIL, while the iCaRL is in a state of balance between forgetting and acquiring knowledge. And the fine-tuning method is forgetting knowledge constantly. Furthermore, by comparing the final classification results of SSCIL and self-supervised joint-learning (black star), we can find that the gap between them is still apparent. As the training phases rise, the gap between them is enlarged: on 5-phase ImageNet-100, the final gap between SSCIL and joint-learning is 10\%, but on 10-phase, the gap has become 15\%. That suggests that the catastrophic forgetting in the SSCIL becomes serious like other supervised methods when the training phases go up.

\begin{figure*}[t]
 \vspace{-.6em}

	\begin{subfigure}[t]{0.33\linewidth}
	\includegraphics[width=2.28in,height=1.5in]{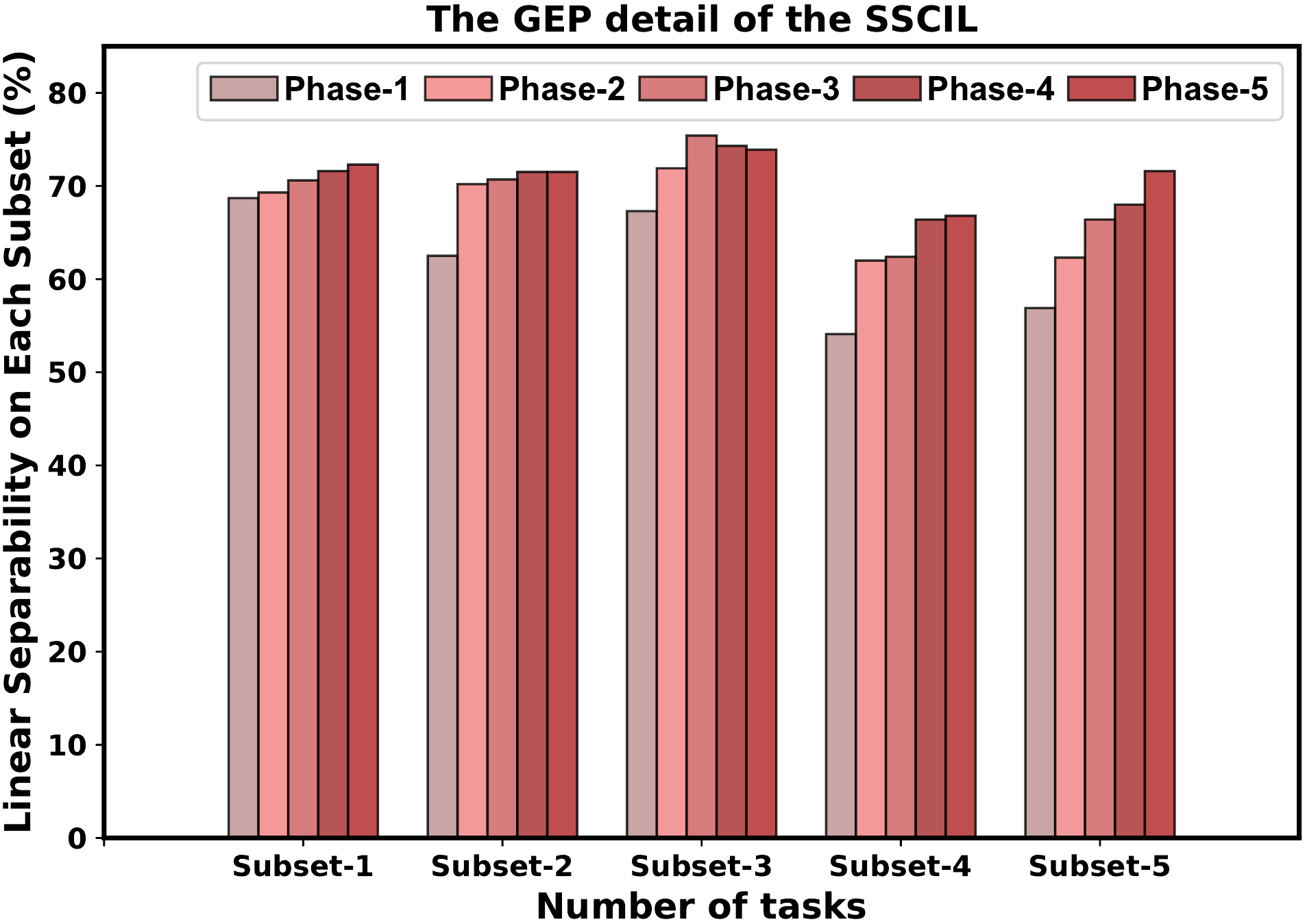}
	\end{subfigure}%
	\begin{subfigure}[t]{0.33\linewidth}
	\includegraphics[width=2.28in,height=1.5in]{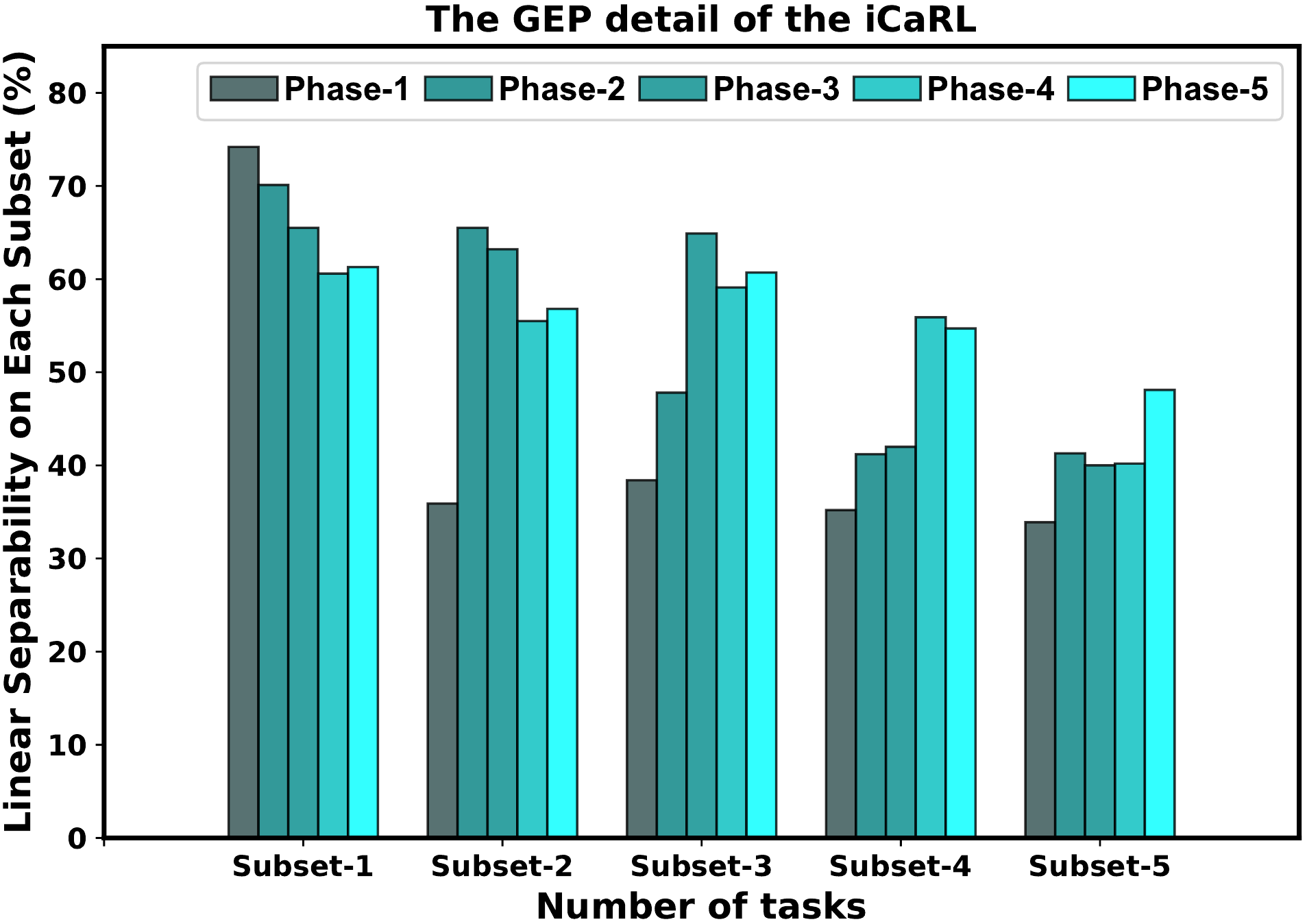}
	\end{subfigure}%
	\begin{subfigure}[t]{0.33\linewidth}
	\includegraphics[width=2.28in,height=1.5in]{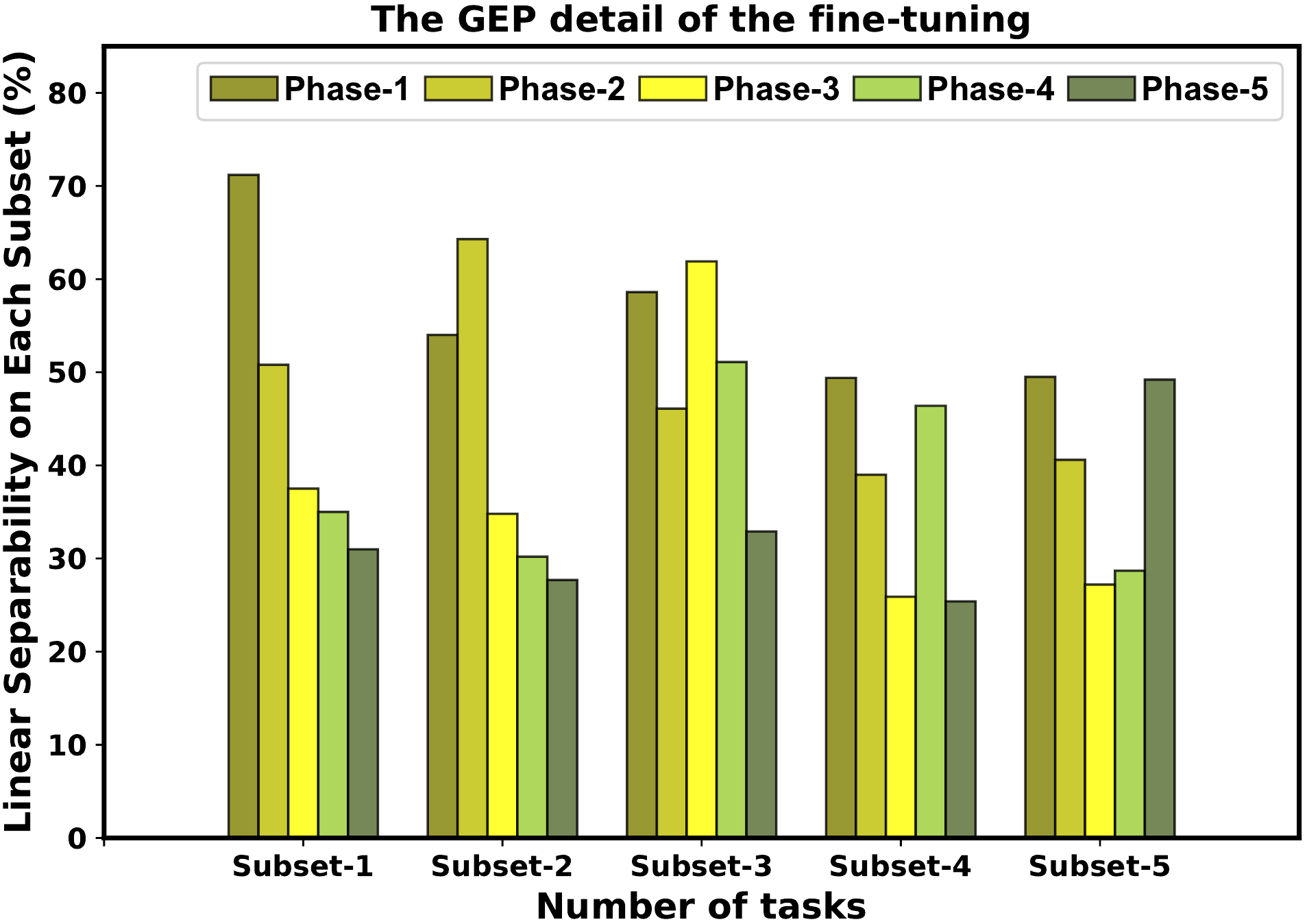}
	\end{subfigure}%
	\vspace{-1em}
	\caption{The GEP detail of the three CIL methods at each phase on 5-phase Imagenet-100 with Random Class Scheme. The horizontal axis represents the  sub-dataset of each phase. Different colors indicate the different GEP results at different training phases.}
	\label{detail_result}
\end{figure*}

\subsection{Result of Different Class Incremental Schemes.}

In order to compare the robustness of SSCIL, iCaRL, and supervised fine-tuning methods, we analyze their performance with three different class incremental schemes, which simulate different class incremental scenarios. From the experimental results of Figure \ref{main_result}, we can see that the performance of all the methods with Random Class Scheme is higher than that of Semantic Class Scheme. However, the difference is that, especially in multi-phase CIL (10-phase), when training in different class incremental sequences, the performance of the SSCIL is more stable than the supervised CIL methods: on ImageNet-100, the final LEP of 10-phase SSCIL is 65\% with Random Class Scheme and 63\% with Semantic Class Scheme, but the result of iCaRL drops from 52\% to 40\%. In addition, comparing the GEP (dotted line) of the SSCIL and iCaRL, we find that the generalization of the models with the Random Class Scheme improves during the CIL. However, in the Semantic Class Scheme, due to the lack of high-level semantic similarity between the sub-datasets, the generalization of iCaRL and fine-tuning fluctuates obviously. Only the GEP of the SSCIL is still rising steadily. All of this shows that SSCIL has a more robust anti-forgetting ability and stability, which ignoring the shift of the data domain in the CIL. 


\subsection{Showing the GEP Details of the Model}
When the GEP of the model increases, it only indicates that the model's generalization is being better. Still, it cannot illustrate whether its ability to classify the old class data has declined. For example, when the model's classification accuracy in the old classes drops by 5\% while the accuracy in the new class data is 7\% higher than the average result, the GEP of the model is still rising. However, in fact, the model forgets its ability to recognize old class data slowly. 



In order to show the details of the model's GEP result at each training phase, we subdivide the model's GEP results into each sub-dataset. Figure \ref{detail_result} shows the detailed results of three methods' GEP at each training phase on 5-phase Imagenet-100 with Random Class Scheme. The horizontal axis corresponds to the sub-dataset of each training phase, and the bar with different color indicate the model's result at the current sub-dataset with varying training phases. From Figure \ref{detail_result}, we find that with the increase of the training phase, the classification accuracy of the SSCIL model on each sub-dataset is improved. It shows that when the SSCIL learning to distinguish the new classes, the model doesn't forget the classification ability for the old classes. The increased performance of the SSCIL on old classes also indicates that the new knowledge acquired from new classes in SSCIL also helps the model recognize the old classes better. On the contrary, the ability of the supervised fine-tuning method to classify old classes decreases significantly with the rise of the training phases. Only the classification accuracy on the  current subset has improved. However, with the training of the new class dataset, its linear separability of the model's representation in previous sub-datasets has declined again. This phenomenon shows that the fine-tuning method cannot retain the knowledge learned in the previous phases. Although the iCaRL method reduces the forgetting of knowledge, there still is a significant decline in its ability to recognize old classes in CIL. 


The results of comparing the performance of the SSCIL, iCaRL, and fine-tuning on the ImageNet-100 and ImageNet prove the advantages of self-supervised representation learning in CIL. Less catastrophic forgetting, stronger generalization ability, and robustness. However, comparing the model's final accuracy between the SSCIL and the joint-learning, there is still a significant gap (10\% on 5-phase and 15\% on 10-phase). Reducing the representation forgetting of the model to shorten the gap between SSCIL and joint-learning will become a new direction in CIL.

\section{Conclusion}
We use three class incremental schemes to explore the performance of self-supervised representation learning on class incremental learning (SSCIL) and proposes Linear Evaluation Protocol (LEP) and Generalization Evaluation Protocol (GEP) to evaluate the quality of the model's representation at each training phase in SSCIL. Finally, we provide a baseline of SSCIL on large-scale datasets Imagenet-100 and Imagenet. In addition, by systematically analyzing the influence of data augmentation and projector in SSCIL, we show that SSCIL is more resistant to forgetting than supervised CIL methods, and the model's generalization is also better.


\textbf{Societal Impact}: this work contributes to fundamental research without any societal impact.

\textbf{Limitations}: our exploration is limited by computational resources and cannot carry out other larger-scale verification experiments.

{\small
\newpage
\bibliographystyle{ieee_fullname}
\bibliography{egbib}
}

\end{document}